%% file: main.tex
\documentclass[]{sjtu}

\usepackage{wrapfig}
\usepackage{tabularx}
\usepackage{textcomp}
\usepackage{stfloats}
\usepackage{url}
\usepackage{verbatim}
\usepackage{graphicx}
\usepackage{titlesec}
\usepackage{tocloft}
\usepackage{adjustbox}
\usepackage{multirow}
\usepackage{pifont}
\usepackage{tikz}
\usepackage{comment}
\usepackage{amsmath,amssymb} 
\usepackage{colortbl}  
\usepackage{color}
\usepackage{booktabs} 
\usepackage{hyperref}
\usepackage{graphicx}    
\usepackage{subcaption} 
\usepackage{booktabs}
\usepackage{amsmath} 
\usepackage{multirow} 
\usepackage{booktabs} 
\usepackage{subcaption} 
\RequirePackage{xspace}
\makeatletter
\DeclareRobustCommand\onedot{\futurelet\@let@token\@onedot}
\def\@onedot{\ifx\@let@token.\else.\null\fi\xspace}

\makeatother

\usepackage{makecell}

\definecolor{adptorange}{RGB}{248, 205, 172}
\definecolor{cmpblue}{RGB}{189, 215, 238}
\definecolor{cmpblue}{RGB}{189, 215, 238}

\definecolor{our_red}{RGB}{232,157,160}
\definecolor{our_blue}{RGB}{136,206,230}
\definecolor{our_orange}{RGB}{246,200,168}
\definecolor{our_green}{RGB}{178,211,164}

\definecolor{attn_code0}{RGB}{247,215,200}
\definecolor{attn_code1}{RGB}{238,169,139}
\definecolor{mlp_code0}{RGB}{204,201,221}
\definecolor{mlp_code1}{RGB}{102,95,153}

\definecolor{token_blue}{RGB}{84, 120, 140}

\definecolor{myMagenta}{rgb}{0.9,0,0.4}

\usepackage{pifont}       
\usepackage{bbding}       
\usepackage{fontawesome}
\usepackage{xspace}

\usepackage{float}

\newlength\savewidth

\newcolumntype{x}[1]{>{\centering\arraybackslash}p{#1pt}}
\newcolumntype{y}[1]{>{\raggedright\arraybackslash}p{#1pt}}
\newcolumntype{z}[1]{>{\raggedleft\arraybackslash}p{#1pt}}

\renewcommand{\paragraph}[1]{\vspace{1mm}\noindent\textbf{#1}}
\usepackage{colortbl}
\usepackage{xcolor}
\usepackage{wrapfig}

\renewcommand{\paragraph}[1]{\vspace{1.25mm}\noindent\textbf{#1}}

\usepackage{algorithm}
\usepackage{listings}

\definecolor{codeblue}{rgb}{0.25, 0.5, 0.5}
\definecolor{codekw}{rgb}{0.35, 0.35, 0.75}
\lstdefinestyle{Pytorch}{
    language = Python,
    backgroundcolor = \color{white},
    basicstyle = \fontsize{9pt}{8pt}\selectfont\ttfamily\bfseries,
    columns = fullflexible,
    aboveskip=1pt,
    belowskip=1pt,
    breaklines = true,
    captionpos = b,
    commentstyle = \color{codeblue},
    keywordstyle = \color{codekw},
}

\definecolor{green}{HTML}{009000}
\definecolor{red}{HTML}{ea4335}

\title{Omni{\color{blue}V}{\color{red}T}{\color{green}L}{\color{orange}A}: {\color{blue}V}ision-{\color{red}T}actile-{\color{green}L}anguage-{\color{orange}A}ction Models with Semantic-Aligned Tactile Sensing}

\author[1\dagger]{Zhengxue Cheng}
\author[2]{Yiqian Zhang}
\author[1]{Anni Tang}
\author[1]{Keyu Wang}
\author[1]{Wenkang Zhang}
\author[2]{Haoyu Li}
\author[2]{Hengdi Zhang}
\author[1]{Li Song}
\affiliation[1]{Shanghai Jiao Tong University}
\affiliation[2]{Paxini Tech.}
\contribution[\dagger]{Corresponding author}

\input{sections/0_abstract}

\date{\today} 
\homepage{https://readerek.github.io/Objtac.github.io/}
\dataset{https://omnisharingdb.paxini.com/}
\correspondence{zxcheng@sjtu.edu.cn}

\begin{document}
\thispagestyle{firstheader}
\maketitle
\pagestyle{empty}

\input{sections/1_introduction}

\input{sections/2_relatedwork}
\input{sections/3_method}

\input{sections/4_experiment}

\input{sections/5_conclusion}

\bibliographystyle{assets/plainnat}
\bibliography{paper}

\newpage
\input{sections/6_supp}


\end{document}

%% file: sections/0_abstract.tex
\abstract{

Recent vision-language-action (VLA) models build upon vision-language foundation models, and have achieved promising results and exhibit the potential of task generalization in robot manipulation. However, due to the difficulty of acquiring tactile data, current VLA models significantly overlook the importance of tactile perception and fail in contact-rich tasks. To address this issue, this paper proposes \textbf{OmniVTLA}, a novel framework that models vision, tactile, and language for end-to-end contact-rich manipulation tasks. Specifically, we introduce \textbf{ObjTac}, a comprehensive tactile dataset comprising textual, visual, and tactile information for 56 objects across 10 categories. Leveraging its 135K tri-modal samples, we train a semantically-aligned tactile vision transformer (SA-ViT) to ground tactile signals in visual and linguistic contexts. Through carefully controlled parameter-matched experiments, OmniVTLA presents a well-designed dual-encoder architecture for tactile perception. Real-world experiments demonstrate substantial improvements over widely-used VLA baselines for two tasks. OmniVTLA achieving 96.9\% success rates with grippers, (21.9\% higher over baseline) and 100\% success rates with dexterous hands (6.2\% higher over baseline) on pick-and-place task and 83.3\% success rates, (33.3\% higher over baseline) on peg-insertion task. Besides, OmniVTLA significantly reduces task completion time and generates smoother trajectories through tactile sensing compared to existing VLA. Our ObjTac dataset can be found at \url{https://readerek.github.io/Objtac.github.io/}.
}

%% file: sections/1_introduction.tex
\section{Introduction} \label{sec:introduction}
Tactile sensing is fundamental to human dexterity, enabling complex tasks, from threading a needle to handling fragile objects—with remarkable precision and adaptability. Although vision provides a global spatial context, tactile sensing offers complementary advantages (e.g., object texture and hardness). It highlights the critical role of vision-tactile integration for manipulation tasks that require physical interaction.
Recent studies on tactile pre-training~\citep{anytouch, tvl, CLTP} have emerged to explore the better representation of tactile data, therefore they usually focus on tactile understanding tasks, and did not necessarily investigate how tactile data changing during interactions and evaluate robotic manipulation tasks.

Recent advances in Vision-Language-Action (VLA) models \citep{brohan2023rt, kim2024openvla, black2024pi_0, team2025gemini} have revolutionized robotic manipulation. They leverage large-scale pretrained vision-language models (VLMs)~\citep{liu2023visual, li2024llava, zhang2025videollama, bai2025qwen2} to interpret natural language instructions and visual observations, demonstrating great potential for generalization and intelligence. However, these models predominantly rely on vision and language, overlooking the rich semantic and physical feedback provided by tactile sensing. 
Existing attempts \citep{vtla, huang2025tactile, yu2025forcevla} to incorporate touch into VLA frameworks often treat tactile data as low-level signals, failing to align them semantically with visual and linguistic contexts.

\begin{figure*}[t!]
    \centering
    \includegraphics[width=1\linewidth]{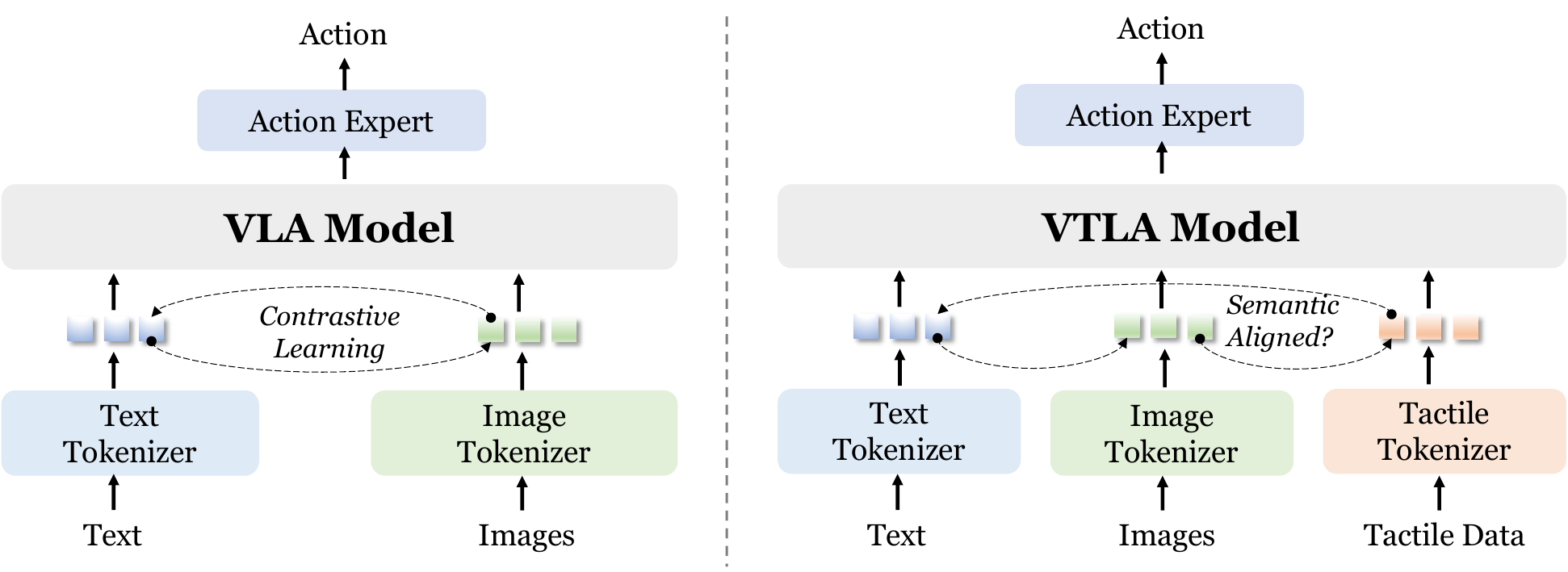}
    \caption{Left: The vanilla VLA model, where the image encoder usually inherits a pretrained CLIP/SigLIP backbone trained with contrastive learning to achieve latent-space semantic alignment. Right: VTLA model. Critically, the design of tactile encoders and semantic alignment among vision, language, and touch modalities is rarely studied.}
    \label{fig:teaser}
\end{figure*}

To fill the gap between tactile pre-training and VLA, we suppose that vision-language-tactile semantic alignment can improve the performance of VTLA by exploring a good choice of tactile encoder architecture. We propose \textbf{OmniVTLA} (Vision-Tactile-Language-Action Model), a novel architecture that unifies vision, touch, and language into a shared semantic space, as shown in Fig.~\ref{fig:teaser}.
OmniVTLA leverages contrastive learning to align high-resolution tactile signals with visual and language concepts, enabling robots to ``understand'' what they feel in the context of what they see and what they are asked to do. Specifically, we introduce a dual-encoder path for tactile data, comprising of a pretrained vision transformer (ViT) and a semantically-aligned tactile ViT (SA-ViT). Second, we build \emph{ObjTac}, a comprehensive dataset capturing textual, visual, and tactile data for 56 objects across 10 categories, a total of 135K tri-modal samples. Third, we train a semantically-aligned tactile encoder using our collected dataset to learn a tactile representation grounding tactile signals (e.g., material, roughness, hardness) in visual and linguistic contexts, serving as a better initialization for OmniVTLA. Extensive experiments demonstrate OmniVTLA’s superiority over VLA baselines. On pick and place tasks, OmniVTLA improves the success rates by 21.9\% to reach 96.9\% with the gripper and improves the success rates by 6.2\% to reach 100\% with the dexterous hand. OmniVTLA attains an 83.3\% success rate on contact-rich peg-insertion tasks, outperforming the baseline by 33.3\%. Moreover, it generates smoother trajectories that adhere to the intuitive principle of \emph{``move quickly when clear, only slow down during contact approach.''}

Our contributions are summarized as follows. 
\begin{itemize}
\item We propose \textbf{OmniVTLA}, a novel framework that models vision, tactile, and language for end-to-end contact-rich manipulation tasks. Through controlled parameter-matched experiments, OmniVTLA presents a well-designed dual-encoder architecture for tactile perception.
\item We introduce \textbf{ObjTac}, a comprehensive tactile dataset, collecting 135K tri-modal samples for 56 objects across 10 categories. Based on it, we train a semantically-aligned tactile encoder for OmniVTLA.
\item Real-world peg-insertion and pick-and-place experiments reveal the superior performance of OmniVTLA compared to typical VLA models, more than 20\% success rate improvement, respectively. Besides, it reduces the completion time and smooths the generated trajectories.
\end{itemize}

%% file: sections/2_relatedwork.tex
\section{Related Works}

The difference of our proposed VTLA and other VLA models are summarized in Table~\ref{tab:diff_models}.

\begin{table}[t]
\footnotesize
\caption{Comparison of different VLA models. L: Language; V: Vision; T: Tactile A: Action.
}
\centering
\setlength{\tabcolsep}{2.0mm}
\begin{tabular}{lccccc}
\toprule
{Model Type}
 & Methods & Input & Output  &Semantic-Aligned    \\
\midrule
VA    & Diffusion Policy \citep{chi2023diffusion}   & V  & A   & \ding{51}  \\
VTA   & RDP \citep{xue2025reactive}    & V + T  & A   & \ding{55}  \\
VLA   & OpenVLA \citep{kim2024openvla}, $\pi0$\citep{black2024pi_0} & V + L & A  & \ding{51} \\
TLA   & TLA \citep{tla}& T + L & A & \ding{55}   \\
VTLA  & \makecell{VTLA \citep{vtla}, \\ Tactile-VLA \citep{huang2025tactile}} & V + T + L & A & \ding{55} \\
OmniVTLA    &Ours &V + T + L  &  A   &  \ding{51} \\
\bottomrule
\label{tab:diff_models}
\end{tabular}
\end{table}

\paragraph{Tactile Sensing for Perception Tasks.}
Early research in tactile sensing focuses on processing low-level physical signals (e.g., force, vibration, deformation) for grasp stability prediction \citep{calandra2018more, cui2020grasp}  and slip detection \citep{li2018slip}, or for dataset construction \citep{tvl, cheng2025touch100k}, shared embedding spaces \citep{unitouch}, transferable architectures \citep{t3}.
Recent works have shifted toward learning general tactile representations for transferability across tasks, sensors, and modalities. Specifically, Octopi~\citep{octopi} is a pioneer work that models physical object properties (i.e., hardness, roughness, and bumpiness) through tactile-grounded understanding, with a primary focus on surface attributes. It significantly improves scenarios reasoning, but overlooks the incorporation into the robotic policy learning.
Anytouch 2~\citep{anytouch2} extends the multi-modal alignment idea proposed in AnyTouch~\citep{anytouch} and contributes a large-scale, five-level tactile dataset covering tactile atomic actions, manipulations, and touch–force paired data. This work further encompasses tactile dynamics during interaction, including force prediction, pose estimation, and slip detection.
In contrast to the above, CLTP~\citep{CLTP} adopts 3D tactile point clouds instead of vision-based tactile representations and aims at 3D contact geometry understanding, such as contact shape, area, position, force, and depth. More recently, FG-CLTP~\citep{FG-CLTP} introduces a fine-grained tactile pretraining method and additionally models the contact states, such as slide, twist, and shear forces.


\paragraph{Vision-Tactile Fusion for Manipulation.}
Recent advances in vision-tactile policy learning have demonstrated remarkable progress in contact-rich manipulation. Reinforcement learning frameworks have effectively combined visual and tactile inputs for assembly tasks \citep{lee2020making, hansen2022visuotactile} and dexterous in-hand manipulation \citep{hu2025dexterous}. More recently, the field has increasingly adopted imitation learning paradigms \citep{yu2023mimictouch, lin2024learning, huang20243d, xue2025reactive, liu2025vitamin}, exploring vision-tactile representations and system architectures for fine-grained manipulation. While achieving impressive task-specific performance, these methods remain limited in semantic reasoning and generalization capabilities compared to vision-language-action models, which remains a large gap that our work would like to address through vision-tactile semantic fusion.

\paragraph{Vision-Language-Action Model.}
Vision-language-action (VLA) models have emerged as a powerful paradigm for generalist robotic policies. \citep{rt2} pioneers this direction by representing robot actions as language tokens, enabling knowledge transfer from web-scale pretraining. \citep{openvla} offers an open-source alternative via LoRA fine-tuning for efficient transfer. Subsequent work \citep{octo, black2024pi_0, rdt-1b, gr00t} expand these capabilities through flow-based or diffusion-based action generation \citep{chi2023diffusion}. Scalability efforts \citep{ wen2025tinyvla, team2025gemini, shukor2025smolvla}), reasoning mechanisms \citep{zhao2025cot, lin2025onetwovla} and 3D extensions \citep{zhen20243d, qu2025spatialvla} further boost applicability. While VLA models excel at open-world generalization, their reliance on vision and language alone limits performance in contact-rich tasks requiring precise physical interaction.

Emerging tactile-enhanced approaches address these limitations through language-based sensor fusion \citep{beyond}, tactile-involved VLA learning \citep{tla, vtla}, and low-dimensional force-aware control \citep{huang2025tactile, yu2025forcevla}.
However, these approaches have not fully explored the design of the tactile encoder. Our OmniVTLA framework fundamentally advances this paradigm by establishing dual-encoder path for touch, through unified cross-modal representation learning.

%% file: sections/3_method.tex
\section{Methods}

\begin{figure*}[t]
    \centering
    \includegraphics[width=1\linewidth]{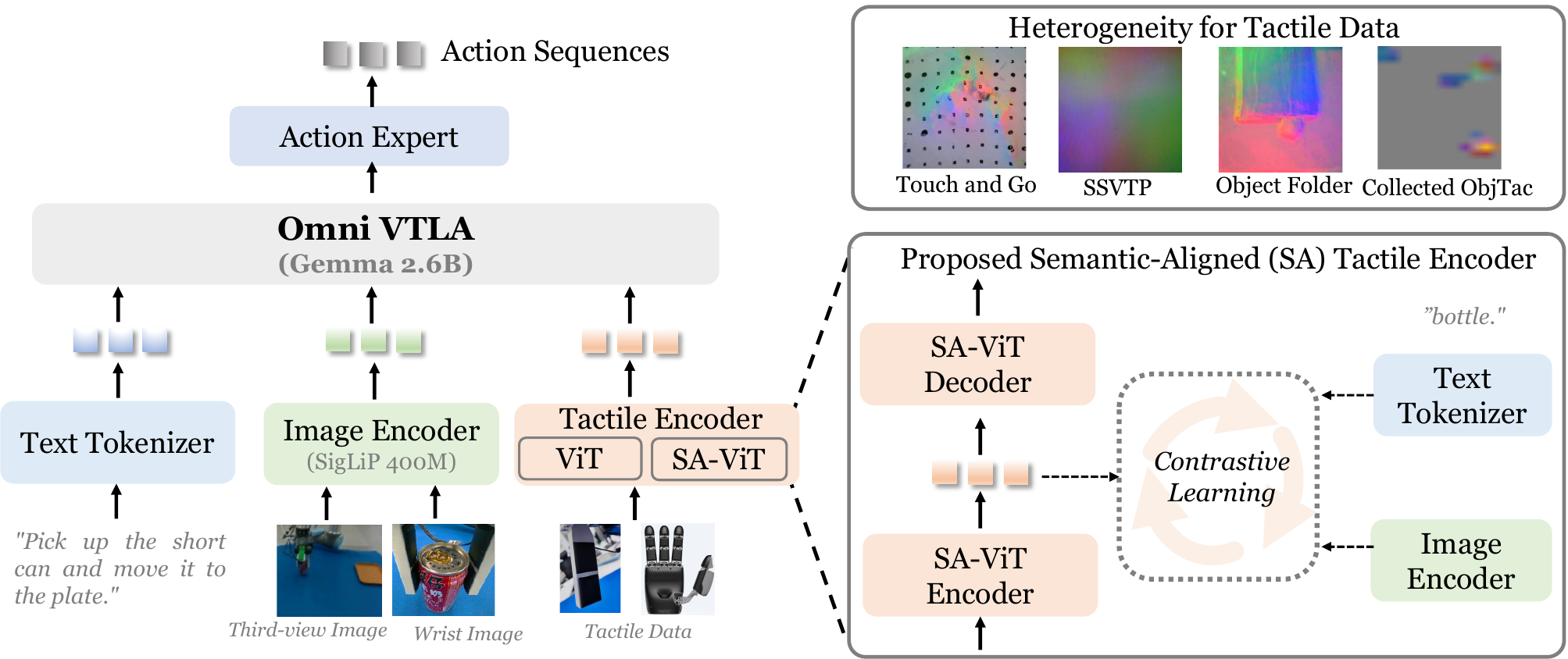}
    \caption{Overview of our proposed \textbf{OmniVTLA}. It integrates dual-ViT encoder for tactile data, to address the inherent heterogeneity between visual and tactile data, as well as across different tactile sensors. The first ViT leverages a pre-trained visual encoder to inherit rich semantic representations from large-scale image data. The second ViT (SA-ViT) is explicitly trained via cross-modal contrastive learning to achieve semantic alignment among tactile, visual, and textual modalities. This dual-encoder design enables effective knowledge transfer and consistent representation learning across diverse sensory inputs.}
    \label{fig:overview}
\end{figure*}

\subsection{Problem Formulation}

Formally, the aim of the action model is to model the distribution $p(\mathbf{A}_t | \mathbf{o}_t)$, where $\mathbf{A}_t = \{a_{t}, a_{t+1}, \dots, a_{t+H-1}\}$ denotes the corresponding sequence of actions (H is the chunk length), and $\mathbf{o}_t$ denotes the observations at the current time. For a typical VLA model, the observation consists of several RGB images, a language prompt, and the robot proprioceptive state, then the model is formally expressed as:
\begin{equation}
    \mathbf{A}_t \sim \mathbf{M}_\text{VLA}(f_\phi(\mathbf{I}_{t}^{i}), l_t),
\end{equation}
where $\mathbf{I}_{t}^{i}$ denotes the $i^{\text{th}}$ image, such as the third-view image and the robot wrist image, $l_t$ is a sequence of language tokens. Usually, images $\mathbf{I}_{t}^{i}$ are encoded using a contrastive image encoder $f_\phi$ (e.g., CLIP, SigLIP) based on Vision Transformers (ViT)~\citep{dosovitskiy2020image} and then projected into the latent embedding space with the text token. 

Meanwhile, our objective of the VTLA model is to incorporate the tactile data into the input, illustrated as Fig.~\ref{fig:overview}. The VTLA model is expressed as follows.
\begin{equation}
    \mathbf{A}_t \sim \mathbf{M}_\text{VTLA}(f_\phi^{I}(\mathbf{I}_{t}^{i}), f_\theta^{T}(\mathbf{T}_{t}^{j}), l_t),
\end{equation}
where $\mathbf{T}_{t}^{j}$ denotes the $j^{\text{th}}$ tactile data, such as the tactile sensor attached to the two-fingertip gripper or the multiple fingers and the palm of the dexterous hand. $f_\phi^{I}$ and $f_\theta^{T}$ denote the image and tactile encoders, respectively. Intuitively, the tactile data can be remapped to a tensor and encoded using the ViT-like structure as an image, but the characteristic of the tactile data is significantly different from the visual data. In this work, our objective is to explore different tactile encoders and corresponding training strategies to validate the best architecture of VTLA.

\begin{figure*}[t]
    \centering
    \includegraphics[width=1.0\linewidth]{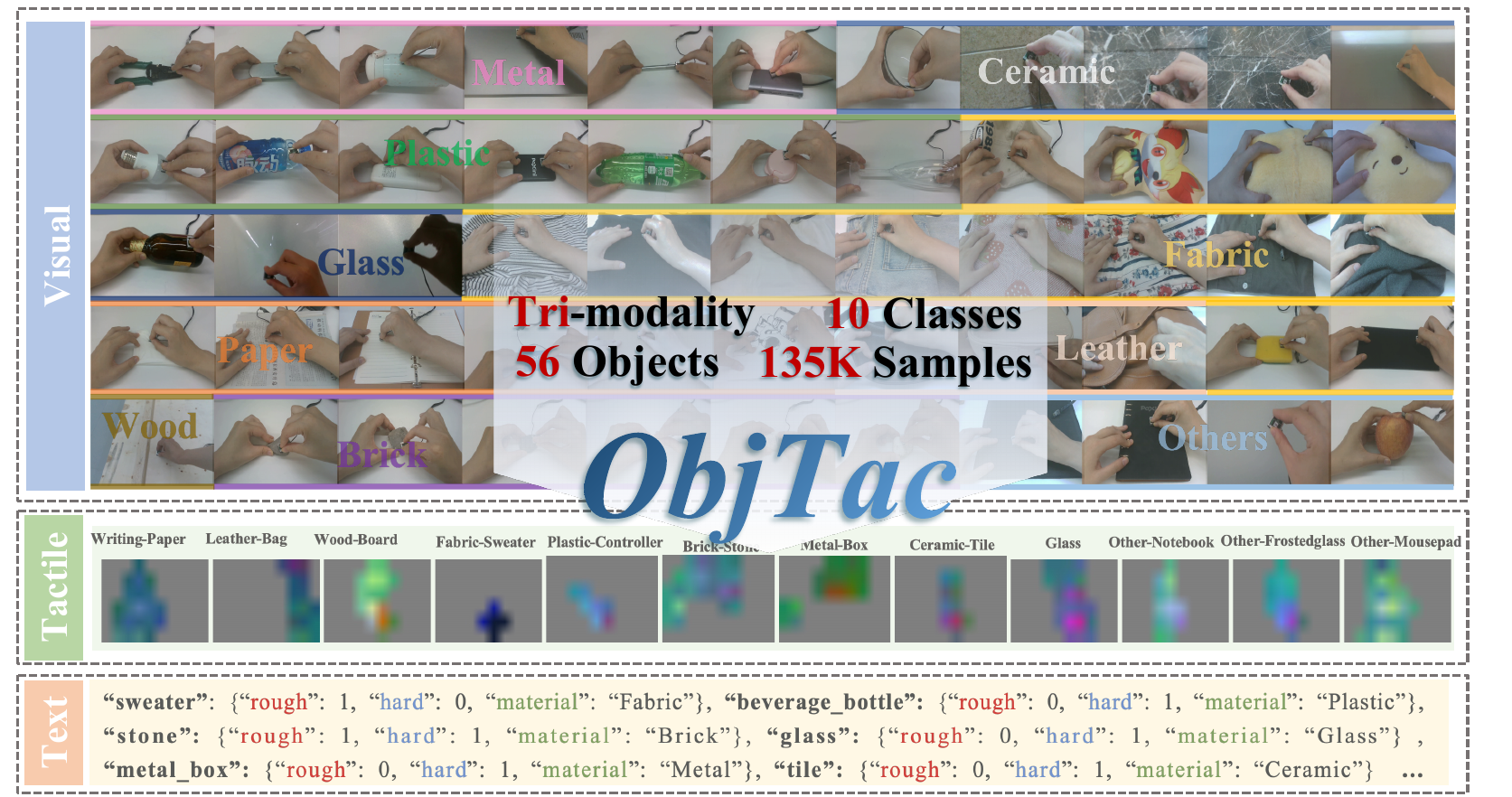}
    \caption{A snapshot of our collected dataset \emph{\textbf{ObjTac}} with 56 objects in ten categories. We collect the visual and tactile data pair to achieve semantic-level alignment. We also visualize part of tactile images after the normalization.}
    \label{fig:dataset}
\end{figure*}

\subsection{Overall Architecture with Dual-Encoder Path}

The proposed OmniVTLA, as shown in Fig.~\ref{fig:overview}, is built on the basis of $\pi0$~\citep{black2024pi_0}. It consists of three core components: tokenizers, backbone, and action head.
The tokenizers process: language instructions $l_t$ via a PaliGemma tokenizer, image encoder  to project $\mathbf{I}{t}^{i}$ using a SigLiP model~\citep{zhai2023sigmoid}, and tactile encoder projecting $\mathbf{T}{t}^{j}$ to latent tokens. 
Specifically, for images including the third-view and wrist, we resize raw captures to $224\times224$, yielding 256 tokens per image. For tactile data, we normalize the data range to int8 and stitch multi-sensor inputs into single images and process resized $224\times224$ inputs through an ViT-like encoder to generate 256 tokens. 
The Gemma-2.6B backbone processes concatenated tokens to produce action tokens, decoded by an action head trained with a flow matching loss following~$\pi0$. Action representations vary by end-effector. For two-finger grippers, they are represented as 10 tokens (3 relative positions, 6 relative angles, 1 gripper state). For four-finger hand, they are 25 tokens (3 relative positions, 6 relative angles, 16 absolute joint positions).

\begin{table*}[t!]
\footnotesize
\centering
\caption{Evaluation of tactile perception capability on Touch and Go dataset, collected by visuo-tactile sensors, and on our collected tactile dataset. \textbf{Bold} font denotes the best performance, and the \underline{underlined} font denotes the second best performance.}
\label{tab:tac-encoder}
\setlength{\tabcolsep}{0.5 mm}
\begin{tabular}{lcc | lll | lll }
\toprule
Encoder & Tactile-Semantic & Our data & \multicolumn{3}{c|}{Touch and Go}  & \multicolumn{3}{c}{Our Collected Dataset}  \\
 & Alignment & in training set & Material & Roughness & Hardness  & Material & Roughness & Hardness  \\   
\midrule
CLIP~\citep{radford2021learning} &\ding{55} &\ding{55} & 54.64 & \underline{86.48} & 86.83 & \underline{54.64} & 70.16 & \underline{91.57} \\
SigLIP~\citep{zhai2023sigmoid} &\ding{55} &\ding{55} & 45.13 & 82.94 & 79.52 & 49.15 & 70.02 & 89.38 \\
Tac-ViT (w/o Align)      &\ding{55} &\ding{55} & 68.94 & \textbf{88.71} & \underline{92.81} & 32.78 & \underline{70.82} & 90.44 \\
AnyTouch~\citep{anytouch} &\ding{51} & \ding{55}    & \textbf{79.39}  & 86.32   & \textbf{95.16}   & 40.21 & 68.01 & 90.11  \\
SA-ViT (Ours)  &\ding{51} & \ding{51}    & \underline{74.90}  & 85.46   & 92.10   & \textbf{70.44} & \textbf{82.21} & \textbf{93.91} \\
\bottomrule
\end{tabular}
\end{table*}

\vspace{-3mm}
\subsection{Semantic-Aligned Tactile Encoder}


To train a semantic-aligned encoder, we use our sensors to collect our own dataset \textbf{ObjTac}, with aligned text, video and tactile data. We collect Tactile-Vision data pairs for 56 distinct objects, as illustrated in Fig.~\ref{fig:dataset}. The resulting dataset comprises 10 object types (i.e., plastic, glass, wood, brick, metal, fabric, leather, ceramic, paper and others), categorized by surface roughness (rough vs. smooth) and material hardness (rigid vs. soft).  
The tactile sensor we used is Paxini Gen2~\citep{PaxiniSensor}). It quantifies contact force fields through a multi-step transduction process. It employs an array of Hall effect sensors~\citep{hall2006} to detect magnetic field variations induced by the displacement of embedded magnetic sources under applied contact forces. Then an embedded inverse problem solver first converts magnetic field changes into 6D pose variations of the magnet array, and these pose estimates are subsequently transformed into multi-dimensional force distributions across the sensor surface via a continuum mechanics inversion algorithm. All computational processes, from raw magnetic field measurements to force signal generation, are executed onboard by an integrated microcontroller unit (MCU). Therefore, we can directly get the force data from these sensors. 


The data collection and processing pipeline is explained below. 
1) For each object, we conducted 2–5 interaction trials, with each trial lasting 10–60 seconds (sampled at 60 Hz). This yielded a total of 270,000 force data recordings. We also capture the first-person-view visual recordings at 720P resolution and 30 FPS, resulting in 252 video sequences with an average duration of 18 seconds. In total, we collect 135K samples with paired tactile and vision data.
2) We added object-level annotations for the language modality, including object name, material type, roughness category, hardness category, video-level metadata, and textual descriptions.
3) Temporal synchronization was performed to align visual and tactile modalities using timestamps.

To train a better semantic-aligned encoder, we add our own collected dataset to existing datasets and employ the second-stage training pipeline of AnyTouch~\citep{anytouch} to achieve semantic alignment. Since our dataset contains tri-modal data pairs, we directly use the total alignment loss as: 
\begin{equation}
\begin{aligned}
        \mathcal{L}_{align}  = 
         \alpha_{VL}*\frac{\mathcal{L}_{V \rightarrow L} + \mathcal{L}_{L \rightarrow V}}{2} +  
         \alpha_{VT}*\frac{\mathcal{L}_{V \rightarrow T} + \mathcal{L}_{T \rightarrow V}}{2} + 
         \alpha_{TL}*\frac{\mathcal{L}_{T \rightarrow L} + \mathcal{L}_{L \rightarrow T}}{2}
\end{aligned}
\end{equation}
where $\mathcal{L}_{V \rightarrow L}$ denotes the loss from vision to language within one batch. $\alpha_{VL}$, $\alpha_{VT}$, $\alpha_{TL}$ are hyper-parameters. Besides, the cross-sensor matching loss with binary cross entropy is also added to the total loss. By incorporating our dataset \textbf{ObjTac}, this semantic-aligned tactile encoder can better adapt to implemented tactile sensors and align semantic representations, grounding tactile signals (e.g., material, roughness, hardness) in visual and linguistic contexts. 

To compare the performance, we list five encoders, i.e. CLIP~\citep{radford2021learning}, SigLIP~\citep{zhai2023sigmoid} and Tactile ViT without alignment objective, i.e. Tac-ViT (w/o Align), AnyTouch and SA-ViT. Here, CLIP~\citep{radford2021learning} and SigLIP~\citep{zhai2023sigmoid} are pretrained encoders on a large scale of vision data instread of tactile data, to compare with SA-ViT with the same parameter count. ViT without alignment loss follows the first training stage of AnyTouch~\citep{anytouch} using masked autoencoder to model the characteristics of tactile data, but removes the second-stage training with cross-modality and cross-sensor semantic alignment. By re-training the classification head, the evaluation of tactile perception capability is listed in Table~\ref{tab:tac-encoder}. CLIP and SigLip exhibit comparatively weaker performance, particularly on material and hardness classification tasks. This suggests inherent differences in the characteristics of visual data versus visuo-tactile data. Furthermore, comparing Tac-ViT (w/o Align) against AnyTouch encoder reveals that incorporating semantic alignment constraints improves material and hardness classification accuracy while incurring only minimal degradation in roughness estimation. This observation aligns with the findings reported in~\citep{anytouch}. Besides, SA-ViT achieves significantly higher classification accuracy on our collected tactile datasets while maintaining near baseline performance on Touch and Go dataset.

%% file: sections/4_experiment.tex
\section{Experiments}


\begin{figure}[t!]
    \centering
    \includegraphics[width=0.8\linewidth]{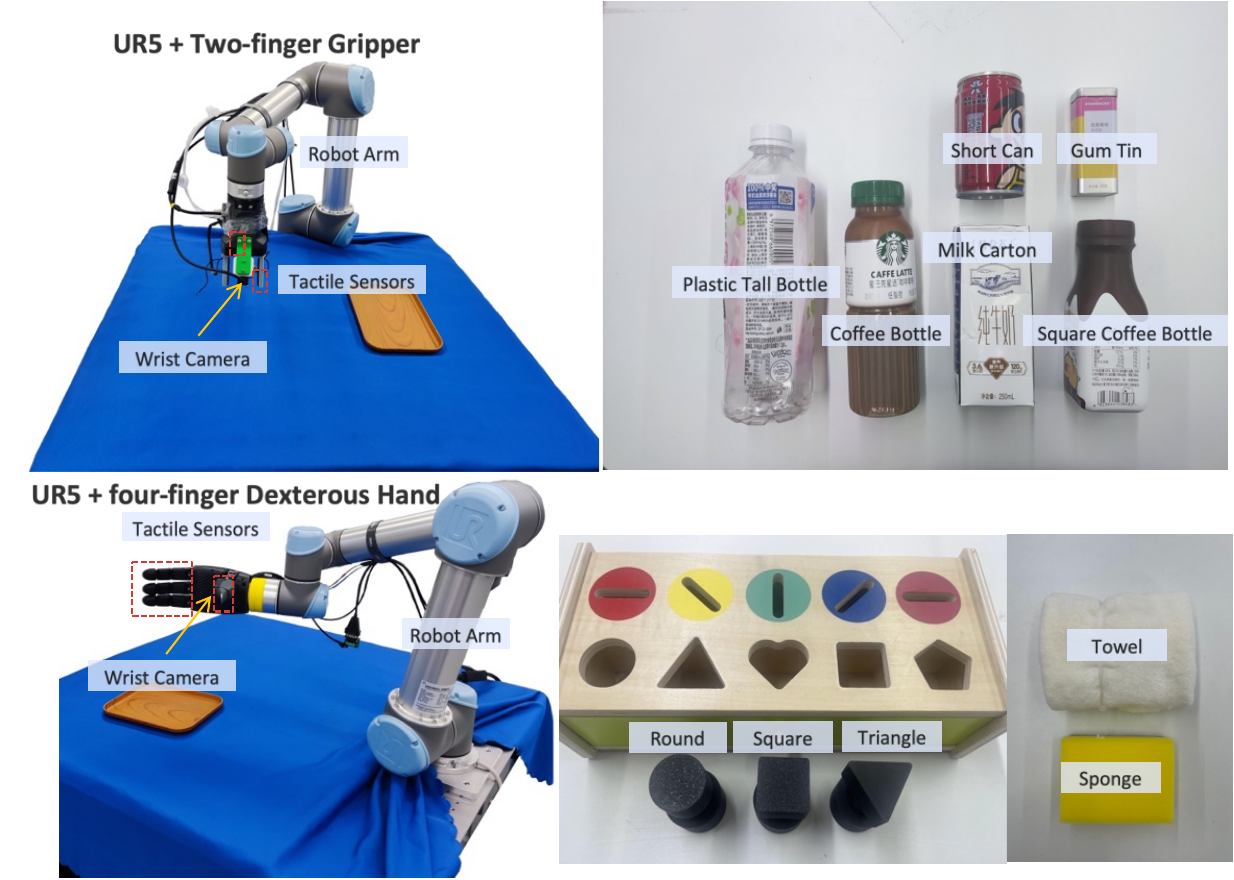}
    \caption{\textbf{Left:} Experimental setup of hardware and environment. \textbf{Right:} Objects involved in manipulation tasks.}
    \label{fig:setup}
\end{figure}

\subsection{Experimental Setup}

\paragraph{Implementation Details} 
Our robotic platform consists of a UR5 arm, a jaw gripper with two tactile sensors and a wrist camera, an 11-tactile-sensor dexterous hand with a wrist camera, and a base camera (Fig.~\ref{fig:setup}). Tactile observations are converted into a 3-channel image representation via max-min force normalization and tensor reshaping. We compare our VTLA model against Diffusion Policy (DP)~\citep{chi2023diffusion} and $\pi0$~\citep{black2024pi_0}. Both baselines are trained using the default configurations from their official codebases, except that DP uses an action chunk size of 64. More training details can be found in the appendix.

\paragraph{Pick-and-place Task Settings} 
We evaluate four objects (\emph{Short Can}, \emph{Square Coffee Bottle}, \emph{Gum Tin}, \emph{Milk Carton}) using the gripper, and two objects (\emph{Coffee Bottle}, \emph{Milk Carton}) using the dexterous hand. We collect 40 teleoperated demonstrations per object at 30 Hz. The plastic bottle and square bottle are treated as unseen objects for generalization evaluation. \emph{Sponge} and \emph{Towel} are unseen objects to evaluate the performance on soft objects. The teleoperated demonstration for the gripper is allowed up to three grasp retries before lifting and transporting the object to a predefined target location. Evaluations are conducted over 32 rollouts for the gripper and 16 rollouts for the dexterous hand using grid-based initial poses. Each trial has a maximum horizon of 1500 steps. 

\paragraph{Peg-insertion Task Settings} 
We conduct tasks by inserting \emph{square}, \emph{round}, and \emph{triangular} blocks into corresponding holes using the gripper with clearances of 0.9 mm, 1.2 mm, and 1.8 mm, respectively. The object is initialized with varying orientations, grasped, moved above the target hole, and inserted through multiple retry attempts. We collect 54 teleoperated demonstrations for \emph{square} and \emph{triangular} blocks at 30 Hz, while the \emph{round} block is reserved for unseen-object generalization evaluation. Each object is evaluated over 10 trials with a maximum horizon of 1500 steps.

\paragraph{Evaluation Variants}
To fully explore a thoughtful design of tactile encoders, we list different settings: 
\begin{itemize}
    \item \textbf{VTLA-FS}: the tactile encoder is trained from scratch and only relies on limited teleoperated tactile data;
    \item \textbf{VTLA-Pre}: the tactile encoder is initialized from a pre-trained vision encoder from a large-scale dataset and fine-tuned on a few teleoperated data;
    \item \textbf{VTLA-SA}: the tactile encoder is first trained from cross-modality contrastive learning to achieve semantic-level alignment (Sec.3.3) and then tuning on a few data;
    \item \textbf{VTLA-Pre-Pre}: dual-encoder path, where one path is SigLIP and the other path is CLIP. These two encoders are pre-trained vision encoders without tactile pereception training, while having the same parameter count as OmniVTLA. It is used as one anchor;
    \item \textbf{VTLA-Pre-Tac}: dual-encoder path, where one path is SigLIP and the other path is Tac-ViT (w/o Align). The only difference from OmniVTLA is without tactile-visual alignment objective; 
    \item \textbf{OmniVTLA}: dual-encoder path, where one path is SigLIP, and the other path is SA-ViT.  
\end{itemize}


\paragraph{Evaluation Metrics}
We evaluated our method through two complementary approaches: \textit{offline validation} and \textit{real-world experiments}. For offline validation , we compute the mean squared error (MSE) between offline-predicted states and ground-truth teleoperation data: $\text{MSE} = \frac{1}{T} \sum_{t=1}^{T} \| x_t - \hat{x}_t \|^2\; $, where $T$ denotes the total timestep and $x_t$ (ground truth) and $\hat{x}_t$ (prediction) represent 10-dimensional or 25-dimensional state vectors comprising end-effector position (xyz), 6D rotation representation \citep{rotation6d}, and 1 gripper aperture or 16 absolute joints for the dexterous hand. 
For real-world evaluation, we employ four metrics: 
\begin{itemize}
    \item (1) \textit{Success Rate} (SR) measuring successful object placement at the end timestamp,
    \item (2) \textit{Completion Time} (CT) from task initiation to successful placement with gripper opening,
    \item (3) \textit{spatial motion smoothness} referring to the end-effector trajectory, quantified by the variability of acceleration along the trajectory,
    \item (4) \textit{motion smoothness of gripper closure}, quantified by the mean absolute changes of the gripper actions along the trajectory. 
\end{itemize}


\begin{figure*}[t!]
    \centering
    \includegraphics[width=1\linewidth]{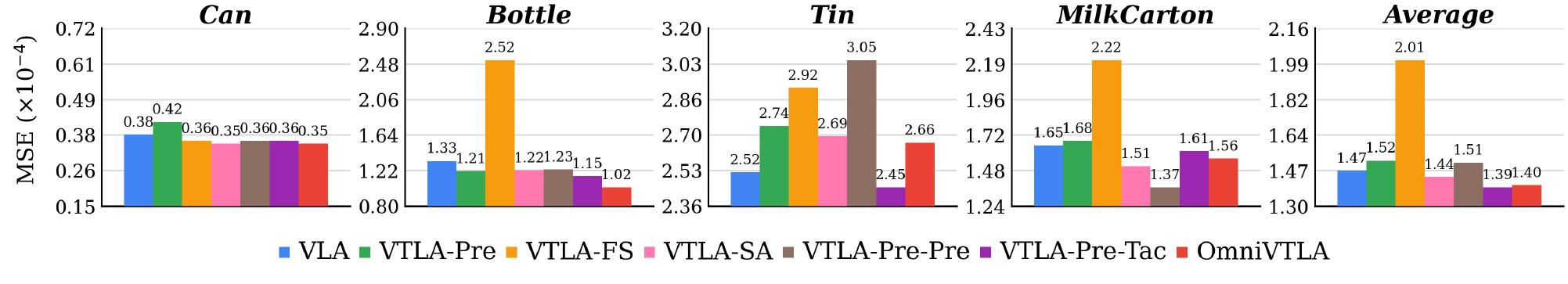}
    \caption{Offline validation results of different models on different objects, where VTLA-Pre-Tac achieves the lowest MSE between predicted trajectory and GT trajectory and OmniVTLA achieves the second lowest MSE on average.}
    \label{fig:val-obj}
\end{figure*}

\begin{table*}[t!]
\footnotesize
\centering
\caption{Real-world pick and place experimental results on different models using two-finger gripper. The baseline VLA model is $\pi0$. \textbf{Bold} font denotes the best performance, and the \underline{underlined} font denotes the second best performance.}
\setlength{\tabcolsep}{2 mm}
\begin{tabular}{cccc|ccccc|ccccc}
\toprule
\textbf{Model} & \multicolumn{3}{c|}{\textbf{Tactile Enc.}} & \multicolumn{5}{c|}{\textbf{SR (\%) $\uparrow$}}  & \multicolumn{5}{c}{\textbf{CT (step) $\downarrow$}}  \\
 & FS & Pre & SA & Can & Bottle & Milk & Tin  & \textbf{Avg} &  Can & Bottle & Milk & Tin  & \textbf{Avg}  \\   
\midrule
VLA ($\pi0$) & \ding{55} &\ding{55} &\ding{55} & 62.5  & 37.5   & 100   & 100    & 75.0  & 981 & 562 & 648 & 436  & 657\\
VTLA-FS  & \ding{51} &\ding{55} &\ding{55}  &75.0  &50.0   &100    &100   &81.2  & 677 & 549 & 498 & 423  &537        \\
VTLA-Pre   & \ding{55} &\ding{51} &\ding{55}  &62.5  &75.0   & 100    &100    &84.4  & 847 & 526 & 540  & 429  &  586 \\
VTLA-SA    & \ding{55} &\ding{55} &\ding{51} &87.5  &62.5  &100    &100  & \underline{87.5} & 524 & 553  & 455 & 405  &\textbf{484}     \\
VTLA-Pre-Pre    & \ding{55} &\ding{51} &\ding{55} &100 & 37.5 & 87.5 & 100 & 81.3 & 618 & 525 & 450 &410 & 501  \\
VTLA-Pre-Tac    & \ding{55} &\ding{51} &\ding{55} &87.5 & 50.0 & 100 & 100 & 84.4 & 665 & 439 & 495 & 602 & 550  \\
\textbf{OmniVTLA}   & \ding{55} &\ding{51} &\ding{51}  &100   &87.5   &100    &100      & \textbf{96.9} & 535 & 537 & 527 & 393  &\underline{498} \\
\bottomrule
\end{tabular}
\label{tab:real-world-pi0-gripper}
\end{table*}

\begin{table*}[t!]
\footnotesize
    \centering
    \caption{Real-world pick and place experimental results on different models using four-finger dexterous hand. The baseline VLA model is $\pi0$. $^\dagger$ denotes the unseen objects, which is not included in training set. $Plastic$ denotes the plastic tall bottle and $Square$ denotes the square coffee bottle. \textbf{Bold} font denotes the best performance.}

    \setlength{\tabcolsep}{2 mm}
        \begin{tabular}{c | ccccc | ccccc }
            \toprule
            \textbf{Model} & \multicolumn{5}{c|}{\textbf{SR (\%) $\uparrow$}}  & \multicolumn{5}{c}{\textbf{CT (step) $\downarrow$}}  \\
             & Bottle & Milk & Plastic$^\dagger$ & Square$^\dagger$  & \textbf{Avg} & Bottle & Milk & Plastic$^\dagger$ & Square$^\dagger$  & \textbf{Avg} \\   
            \midrule
            VLA ($\pi0$)    &100  &100  &87.5  &87.5  &93.8   & 312 & 324  & 369 & 368 &  343 \\
            \textbf{OmniVTLA}   &100  &100  &100  &100  & \textbf{100}
                 & 307 & 305  & 339 & 335  &\textbf{322} \\
            \bottomrule
        \end{tabular}
    \label{tab:real-world-pi0-hand}
\end{table*}

\begin{table*}[t!]
\footnotesize
\centering
\caption{Real-world pick and place experimental results with and without tactile encoders using two-finger gripper. The compared baseline is DP~\citep{chi2023diffusion} and all the parameters are trained from scratch. \textbf{Bold} font denotes the best-performance.}
\setlength{\tabcolsep}{2.0 mm}
\begin{tabular}{cc | lllll | lllll }
    \toprule
    \textbf{Model} & \textbf{Tactile Enc.} & \multicolumn{5}{c|}{\textbf{SR (\%) $\uparrow$}}  & \multicolumn{5}{c}{\textbf{CT (step) $\downarrow$}}  \\
     & & Can & Bottle & Milk & Tin  & \textbf{Avg.} &  Can & Bottle & Milk & Tin  & \textbf{Avg.}  \\   
    \midrule
    VA (DP) & \ding{55}    & 75.0  & 75.0   & 50.0  &  37.5   & 59.4   & 767  &989  & 1010  & 638 & 851 \\
    VTA (Ours) & \ding{51}   & 100  & 75.0  & 75.0  & 62.5  & \textbf{78.1}    & 695 & 658 & 783 & 593 & \textbf{682}    \\
    \bottomrule
\end{tabular}

\label{tab:real-world-dp}
\end{table*}

\begin{table*}[t!]
\footnotesize
    \centering
    \caption{Real-world peg-insertion experimental results on different models using two-finger gripper. $^\dagger$ denotes the unseen shape, which is not included in training set. \textbf{Bold} font denotes the best performance.}
    \label{tab:peg-insertion}
    \setlength{\tabcolsep}{2.4 mm}
    \begin{tabular}{cccc|cccc|cccc}
        \toprule
        \textbf{Model} & \multicolumn{3}{c|}{\textbf{Tactile Enc.}} & \multicolumn{4}{c|}{\textbf{SR (\%) $\uparrow$}}  & \multicolumn{4}{c}{\textbf{CT (step) $\downarrow$}}  \\
                       & FS & Pre & SA & Triangle & Square & Round$^\dagger$ & \textbf{Avg} & Triangle & Square & Round$^\dagger$ & \textbf{Avg} \\   
        \midrule
        VLA ($\pi0$)       & \ding{55} & \ding{55} & \ding{55} & 60.0 & 60.0 & 30.0 & 50.0 & 726 & 873 & 850 & 816 \\
        VTLA-FS            & \ding{51} & \ding{55} & \ding{55} & 60.0 & 40.0 & 40.0 & 46.7 & 606 & 613 & 970 & 730 \\
        VTLA-Pre           & \ding{55} & \ding{51} & \ding{55} & 70.0 & 60.0 & 80.0 & 70.0 & 742 & 683 & 1294 & 906 \\
        VTLA-SA            & \ding{55} & \ding{55} & \ding{51} & 60.0 & 70.0 & 80.0 & 70.0 & 620 & 673 & 740 & \textbf{678} \\
         VTLA-Pre-Pre            
        & \ding{55} & \ding{51} & \ding{55} 
        & 40.0 & 50.0 & 70.0 & 53.3 
        & 640 & 676 & 1081 & 799 
        \\
         VTLA-Pre-Tac            
        & \ding{55} & \ding{51} & \ding{55} 
        & 40.0 & 40.0 & 50.0 & 43.3 
        & 919 & 697 & 1114 &  910 
        \\
        
        \textbf{OmniVTLA}  & \ding{55} & \ding{51} & \ding{51} &70.0  & 90.0 & 90.0 & \textbf{83.3} &765 &583 &1058 & 802 \\
        \bottomrule
    \end{tabular}
\end{table*}

\begin{table*}[t!]
\footnotesize
\centering
\caption{The smoothness of generated trajectories with difference tactile encoders. \textbf{Bold} font denotes the best performance, and the \underline{underlined} font denotes the second best performance.}
\label{tab:smoothness}
\setlength{\tabcolsep}{1.8 mm}
    \begin{tabular}{cccc | ccccc | ccccc }
    \toprule
    \textbf{Model} & \multicolumn{3}{c|}{\textbf{Tactile Enc.}}
     & \multicolumn{5}{c}{\textbf{Smoothness of End-effector ($\times 10^{-6}$) }$\downarrow$ } 
     & \multicolumn{5}{|c}{\textbf{Smoothness of Gripper ($\times 10^{-3}$) }$\downarrow$ } \\
     &FS & Pre & SA & Can & Bottle & Milk & Tin & \textbf{Avg}  & Can & Bottle & Milk & Tin & \textbf{Avg}  \\   
    \midrule
    VLA ($\pi0$) & \ding{55} &\ding{55} &\ding{55}  &2.60 &2.22 &3.54 &2.46 &2.70 &1.98 & 2.26 &2.83 & 3.02 & 2.52\\
    VTLA-FS     & \ding{51} &\ding{55} &\ding{55}   &3.05 &2.39 &3.12 &3.48 &3.01 &1.73 & 1.98 &2.78 & 3.06 & 2.39\\
    VTLA-Pre     & \ding{55} &\ding{51} &\ding{55}  &2.13 &2.20 &3.54 &3.38 &2.81 &1.61 & 2.00 &2.39 & 3.26 & 2.31 \\
    VTLA-SA     & \ding{55} &\ding{55} &\ding{51}   &1.91 &2.48 &2.40 &2.97 &\textbf{2.44} &1.52 & 1.89 &2.21 & 3.04 & \textbf{2.17} \\
{ VTLA-Pre-Pre}    & \ding{55} &\ding{51} &\ding{55}  &2.50 &2.83 &3.29 &3.44 &3.02 &1.67 & 2.07 &2.87 & 3.00 & 2.40 \\
{ VTLA-Pre-Tac}    & \ding{55} &\ding{51} &\ding{55} &2.29 &2.42 &3.79 &1.79 &\underline{2.57} &1.51 & 2.01 &2.53 & 3.95 & 2.50\\
    \textbf{OmniVTLA}     & \ding{55} &\ding{51} &\ding{51} &3.12 &2.86 &2.13 &2.48 &2.65 &1.53 & 1.92 &2.56 & 3.04 & \underline{2.26} \\
    \bottomrule
    \end{tabular}
\end{table*}

\subsection{Evaluation Results}

\paragraph{Validation Results}
Offline validation on teleoperation-driven validation data demonstrates the superior predictive performance of OmniVTLA across diverse objects. As illustrated in Fig.~\ref{fig:val-obj} , VTLA-Pre-Tac achieves the lowest MSE between predicted trajectory and GT trajectory. OmniVTLA achieves the second lowest MSE. This trend holds across most objects: Abnormal results for VTLA-FS might result from the overfitting, demonstrate the importance to use large-scale tactile data, instead of only teleoperated-driven data.

\paragraph{Real-World Pick and Place Results: }
For $\pi0$ using the gripper (Table~\ref{tab:real-world-pi0-gripper}), VTLA-SA outperforms other designs with up to one tactile decoder. It hits an 87.5\% average SR, 6.3\% higher than VTLA-FS and 3.1\% above VTLA-Pre. Our OmniVTLA achieves the best performance, with an outstanding 96.9\% average SR. 
In terms of CT, VTLA-SA reduces the average step count by 26.3\% compared to VLA (from 657 to 484 steps), proving that tactile feedback optimizes manipulation. OmniVTLA achieves the second best performance, reducing the CT by 24.2\% (from 657 to 498 steps).
Our results also demonstrate that, under identical parameter settings, VTLA-Pre-Pre and VTLA-Pre-Tac underperform compared to OmniVTLA in terms of SR, CT and and smoothness. This indicates the performance gains are attributed to the choice of tactile encoder architecture rather than the parameter count. 

For $\pi0$ using the four-finger dexterous hand (Table~\ref{tab:real-world-pi0-hand}), our OmniVTLA increases the SR by 6.2\% (from 93.8\% to 100\%) and cuts the CT by 6\% (from 343 to 322 steps). Especially for unseen objects Plastic and Square, ours achieves the SR of 100\%, but VLA only achieves the SR of 87.5\%.

For the DP baseline (Table~\ref{tab:real-world-dp}), integrating tactile sensing boosts the average SR by 18.7\% (from 59.4\% to 78.1\%) and slashes average CT by 19.9\% (from 851 to 682 steps). This confirms that tactile signals universally enhance the performance, regardless of the baseline.

\paragraph{Real-World Peg Insertion Results}
Results in Table~\ref{tab:peg-insertion} also validate OmniVTLA’s superiority over VLA ($\pi0$) for peg-insertion tasks. OmniVTLA achieve 83.3\% SR on average for two in-distribution shapes and one OOD shape, which is 33.3\% higher than VLA. Meanwhile, in terms of Completion Time (CT), VTLA-SA achieves the fast CT (from 816 to 678 steps). 
With identical settings, VTLA-Pre-Pre and VTLA-Pre-Tac underperform OmniVTLA and VTLA-SA in both success rate and completion time, confirming that the performance gains stem from the semantic-aligned tactile encoder. This is likely because the semantic-aligned encoder generalizes better to new tactile data due to its self-supervised learning, whereas Tac-ViT, trained via supervised perception tasks, may generalize less effectively to manipulation.

\subsection{Analysis and Discussion}

\paragraph{Smoothness of Trajectories} 
Results are listed in Table~\ref{tab:smoothness}. VTLA-SA achieves the best performance on both smoothness of end-effector and gripper, which demonstrates the effectivness of semantic-aligned tactile sensing. VTLA-Pre-Tac achieves the second best performance for the smoothness of end-effector. Regarding the smoothness of the gripper, VTLA-SA achieves the best performance and OmniVTLA achieves the second best performance to achieve smoother gripper actuation during the contact-rich interaction. 


\begin{table}[t!]
    \footnotesize
    \centering
    \caption{Pick and place results using two-finger gripper on two soft objects. The baseline VLA model is $\pi0$. $^\dagger$ denotes the unseen objects, which are not included in training set. \textbf{Bold} font denotes the best performance, and the \underline{underlined} font denotes the second best performance.}
    \label{tab:pick-soft}
    \setlength{\tabcolsep}{4 mm}
    \begin{tabular}{c ccc|ccc}
        \toprule
        \textbf{Model}  & \multicolumn{3}{c|}{\textbf{SR (\%) $\uparrow$}}   & \multicolumn{3}{c}{\textbf{Max Gripper Width $\downarrow$}} \\
                        & Towel$^\dagger$ & Sponge$^\dagger$ &  \textbf{Avg} 
                        & Towel$^\dagger$ & Sponge$^\dagger$ &  \textbf{Avg}\\   
        \midrule
        VLA ($\pi0$)        & 37.5 & 62.5 & 50.0  & 0.652 & 0.711 & 0.682  \\
        VTLA-FS            & 87.5 & 100.0 & \textbf{93.8} & 0.768 & 0.771 & 0.770  \\
        VTLA-Pre           & 87.5 & 100.0 & \textbf{93.8}  & 0.626 & 0.685 & 0.656  \\
        VTLA-SA            & 50.0 & 75.0 & 62.5 & \textbf{0.549}  & \textbf{0.566} & \textbf{0.558} \\
        VTLA-Pre-Pre            
         & 75.0 & 50.0 & 62.5   &0.747 & 0.722 & 0.734 \\
         VTLA-Pre-Tac 
        & 75.0 & 75.0 & 75.0 & 0.611 & 0.612 & 0.611 \\
        \textbf{OmniVTLA}  
        &75.0 & 87.5 & 81.3 
        & \underline{0.578} & \underline{0.587} & \underline{0.583}  \\
        \bottomrule
    \end{tabular}
\end{table}

\begin{figure}[t!]
    \centering
    \includegraphics[width=0.88\linewidth]{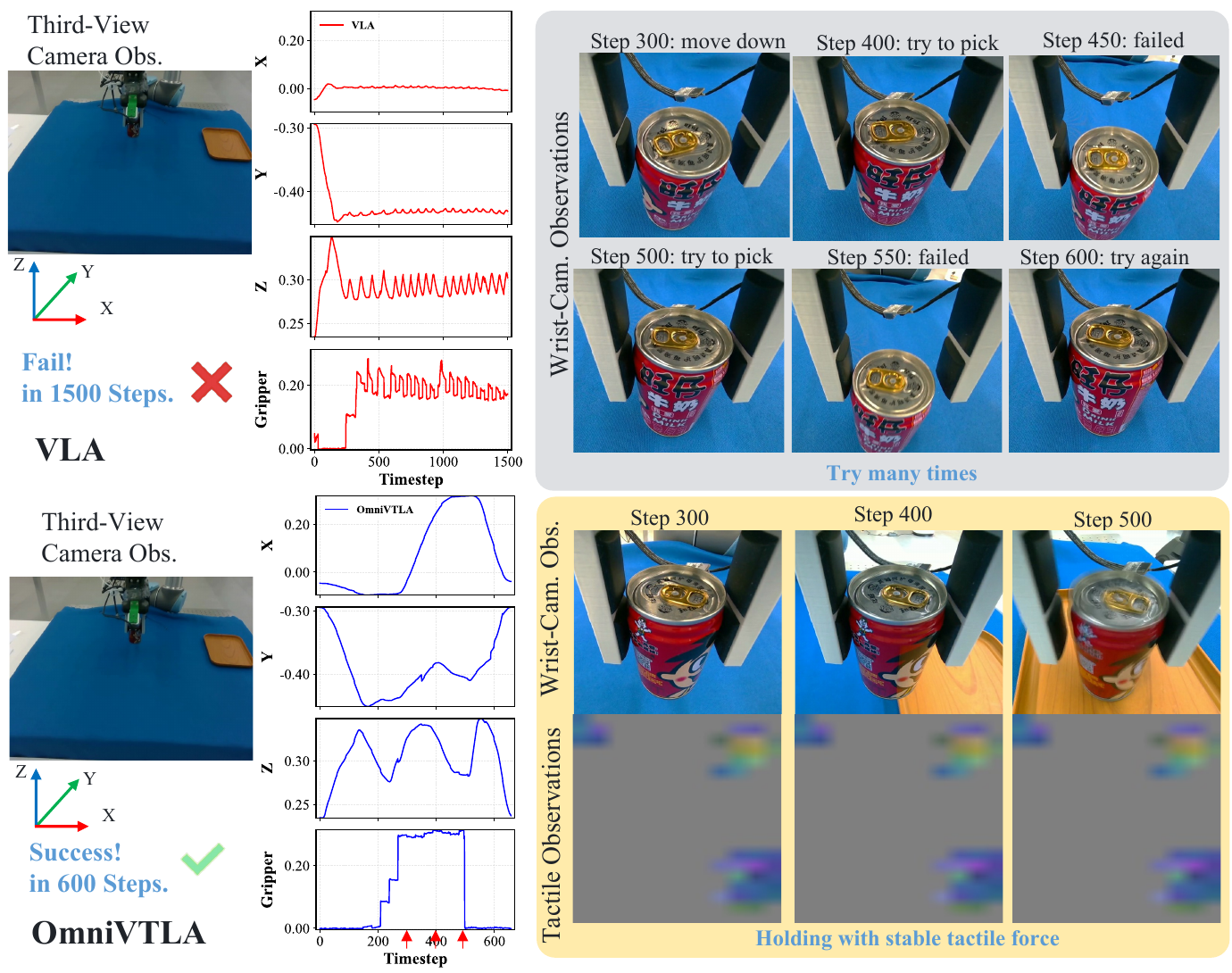}
    \caption{Stage-by-stage visualization for pick and place task, with two models VLA and OmniVTLA.}
    \label{fig:vis-pick-stage}
\end{figure}

\begin{figure}[t!]
    \centering
    \includegraphics[width=0.88\linewidth]{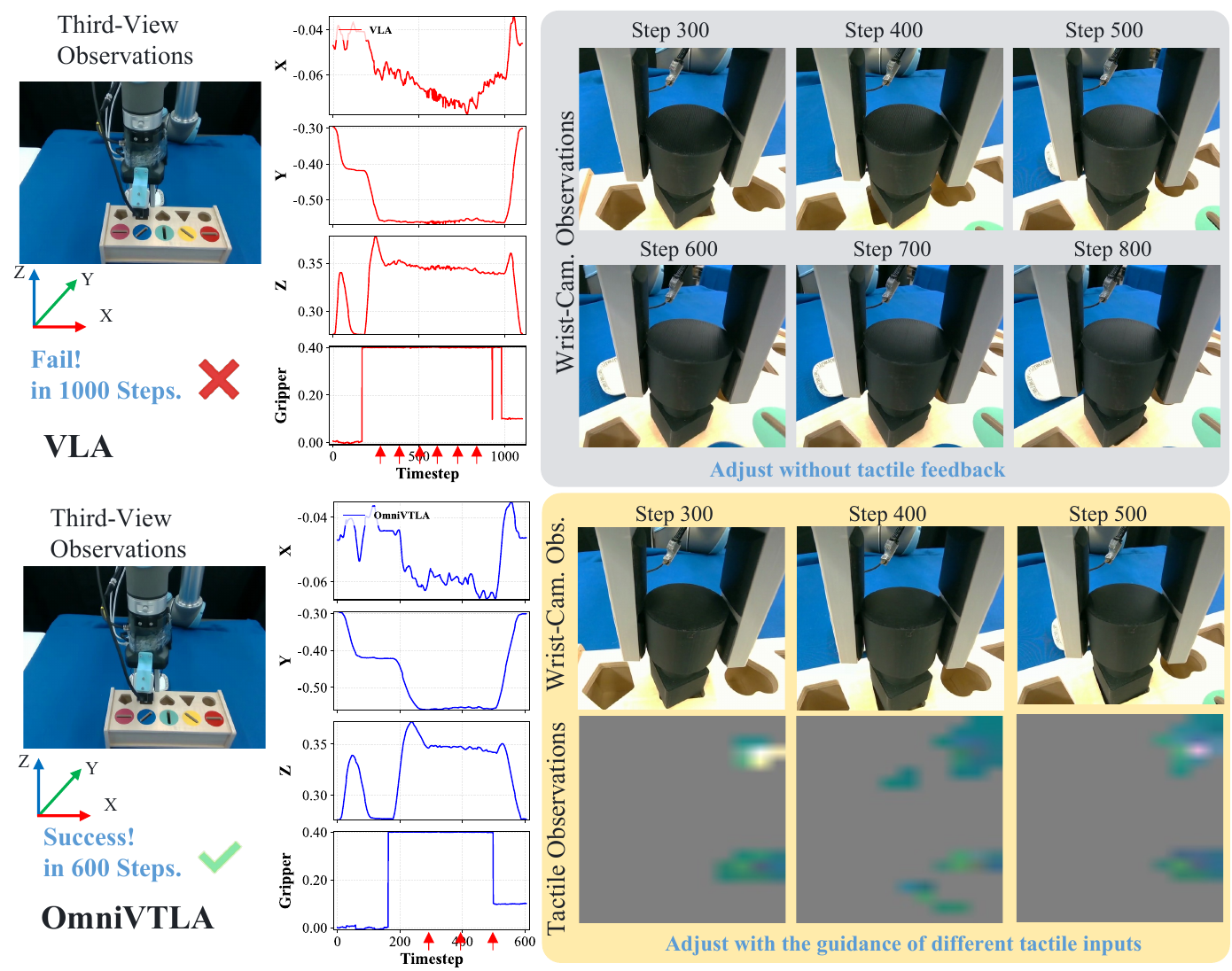}
    \caption{Stage-by-stage visualization for peg-insertion task, with two models VLA and OmniVTLA.}
    \label{fig:vis-insertion-stage}
\end{figure}

\begin{figure}[t!]
    \centering
    \includegraphics[width=1\linewidth]{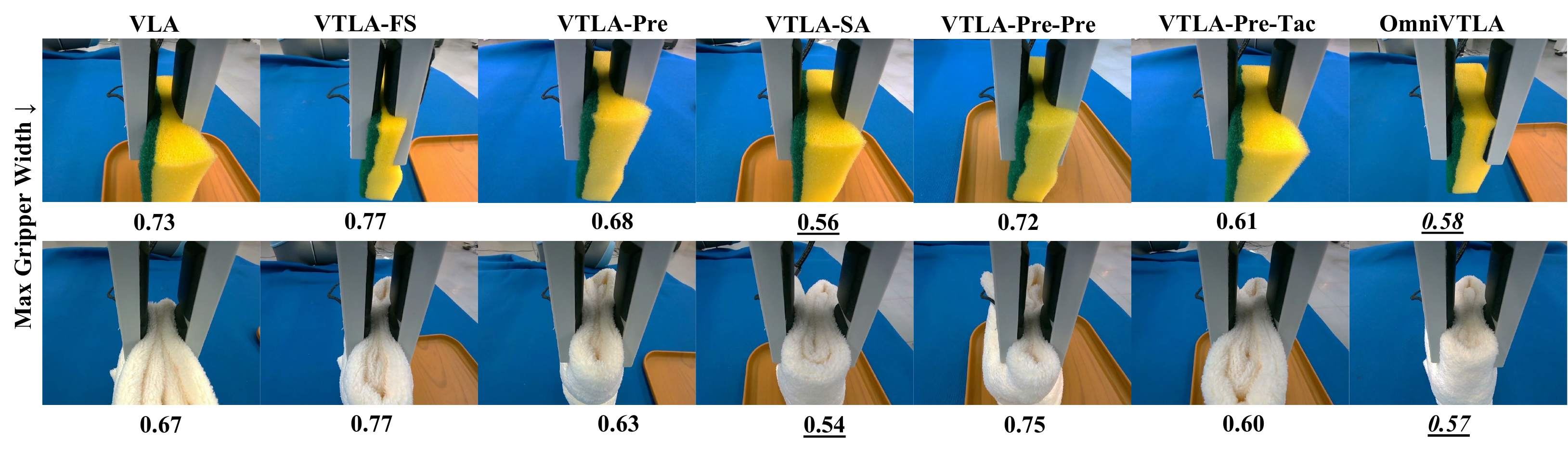}
    \caption{Visualization of maximal gripper width for different models, on Towel and Sponge. \textbf{\underline{{Underlined}}} font denotes the best result, and \textbf{\underline{\textit{italics}}} font denotes the second best result.}
    \label{fig:soft}
\end{figure}

\paragraph{Qualitative Results}
We present stage‑by‑stage visualizations for the pick‑and‑place task and the peg‑insertion task in Fig.~\ref{fig:vis-pick-stage} and Fig.~\ref{fig:vis-insertion-stage}. 
Fig.~\ref{fig:vis-pick-stage} shows VLA repeatedly failing while the lower panel shows OmniVTLA successfully grasping and lifting the bottle with stable tactile force, until releasing. This shows that without tactile input, the policy easily falls into trial‑and‑error loops, whereas tactile input enables stable grasping. Fig.~\ref{fig:vis-insertion-stage} compares failed (VLA) and successful (OmniVTLA) peg insertions under identical initial states. VLA attempts insertion but fails and continues adjusting without success, especially when the peg occludes the hole. OmniVTLA uses tactile feedback to adjust its position and succeeds. This confirms that tactile input is essential for contact‑rich tasks, particularly when vision is occluded.

\paragraph{Generalized to soft objects} 
We further test two unseen soft objects: a \textit{Sponge} and a \textit{Towel}. The results are listed in Table~\ref{tab:pick-soft}. VLA achieves the worst performance. VTLA-FS and VTLA-Pre attain the highest success rate (SR), but this is largely due to very tight gripper closure. In contrast, VTLA-SA and OmniVTLA achieve the top two performances in terms of maximum gripper width, suggesting that the semantic-aligned tactile encoder can recognize soft materials and avoid excessive deformation. Fig.~\ref{fig:soft} visualizes the maximum gripper width, clearly showing the deformation. VTLA-SA and OmniVTLA demonstrate better performance in achieving soft grasping.

%% file: sections/5_conclusion.tex
\section{Conclusion}
We present OmniVTLA, a vision‑tactile‑language‑action model with a semantic‑aligned tactile encoder that leverages visual and language modalities. Our design adopts a dual‑encoder architecture and introduces the ObjTac dataset for cross‑modal contrastive learning, enabling robots to interpret tactile data in task‑relevant contexts. OmniVTLA significantly outperforms state‑of‑the‑art VLA baselines: on pick‑and‑place tasks, it improves success rates by 21.9\% (two‑finger gripper) and 6.2\% (four‑finger dexterous hand); on peg‑insertion, it achieves a 33.3\% higher success rate. It also reduces completion time and produces smoother trajectories through tactile‑guided learning.
Despite these promising results, OmniVTLA still has several limitations. First, our experiments used repeated‑attempt training data for both tasks, but we found that single‑attempt data sufficed for pick‑and‑place; only contact‑rich tasks required multiple attempts. Future work should further explore the role of tactile sensing for general tasks and investigate how to leverage tactile signals without introducing visual ambiguity, aiming for a more generalizable multimodal policy. Second, OmniVTLA does not improve performance in messy clutter, where success depends on visual navigation rather than tactile feedback. Future work will address more complex tasks, more efficient tactile representations, and temporally dynamic fusion architectures.


%% file: sections/6_supp.tex
\section{Appendix}

\subsection{Dataset and Training Details}
\paragraph{Dataset Object List} Table~\ref{tab:item-list} provides the complete object inventory for our \emph{\textbf{ObjTac}} dataset, comprising 56 objects across ten categories.

\begin{table*}[htb]
    \footnotesize
    \centering
    \caption{List of items in our dataset \textbf{ObjTac}, consisting of 10 classes.}
    \setlength{\tabcolsep}{8 mm}
    \begin{tabular}{c|c}
        \toprule
        \textbf{Material} & \textbf{Corresponding Items ($\sim$56)} \\
        \midrule
        Plastic & \makecell[c]{Plastic bulb, Beverage bottle 1,\\  Beverage bottle 2,  Remote control,\\  
        Phone case,  Plastic cup lid,  Plastic goblet} \\
        
        \hline
        Glass & Glass bottle, Glass 1, Glass 2 \\

        \hline
        Wood & Wooden board \\

        \hline
        Brick & \makecell[c]{Stone 1, Stone 2, Stone 3,  Pebble 1, Pebble 2, Pebble 3} \\

        \hline
        Metal & \makecell[c]{Vice, Metal box, Thermos cup,  Laptop, Fountain pen, Adapter} \\

        \hline
        Fabric & \makecell[c]{Pure cotton fabric 1, Pure cotton fabric 2,  Pure cotton fabric 3,\\ Jeans, 
        Pillowcase,  Linen pants, Nylon shirt, Sweater, \\ Sponge 1, Sponge 2, Canvas peaked cap, \\ Plush toy 1, Plush toy 2, Plush toy 3,  Plush toy 4} \\
        
        \hline
        Leather & Leather bag 1, Leather bag 2, Leather bag 3 \\
        \hline
        Ceramic & \makecell[c]{Ceramic bowl, Ceramic tile 1, \\ Ceramic tile 2, Ceramic tile 3,  Ceramic tile 4} \\
        \hline
        Paper & \makecell[c]{Toilet paper, Newspaper, Writing paper, \\ Business card, Corrugated paper,  Paper shopping bag} \\
        \hline
        Others & \makecell[c]{Apple, Frosted glass, Mouse pad,  Notebook cover} \\
        \bottomrule
    \end{tabular}
    
    \label{tab:item-list}
    
\end{table*}

\begin{table*}[tbh]
\footnotesize
\caption{Training details for VTLA, Pi0, VTA and DP.}  
\centering
\begin{tabular}{c | c | c}                                
\toprule
\textbf{Parameter} & \textbf{VTLA \& Pi0} & \textbf{VTA \& DP} \\  
\midrule
GPU                     & NVIDIA A100 (80 VRAM) & NVIDIA A100 (80 VRAM) \\
training method         & fine-tune & train from scratch \\
learning rate           & \makecell[c]{2.5e-5 peak LR \\ (1K steps linear warmup, \\ 29K steps cosine decay to 2.5e-6)} & 0.0001 \\
total batch size        & 32 & 32 \\
train steps             & 30K & 200K \\
input image type            & \makecell[c]{1 third-person camera image, \\ 1 wrist-mounted camera image \\ 1 tactile image (VTLA)}  & \makecell[c]{1 third-person camera image, \\ 1 wrist-mounted camera image \\ 1 tactile image (VTA)} \\
action chunk size       & 50 steps & 64 steps \\
input image size        & 224x224 & 480x640 \\
observation history & no & yes (2-step history) \\
robot state         & yes (use EEF) & yes (use joint) \\
image augmentations & yes & yes \\
\bottomrule
\end{tabular}
\label{tab:hyper_vtla_vta}                            
\end{table*}

\paragraph{Data Collection Process}
Data collection consists of two processes: \emph{Touch} and \emph{Grasp}.

For the \emph{Touch} process, each object is touched 2–5 times, with individual interactions lasting between 10–60 seconds (sampled at 60Hz). A Python script records finger-tactile sensor data alongside precise timestamps, while an Intel RealSense2 camera captures synchronized first-person RGB video at 720p resolution (30 FPS). Across all 56 objects, this process yields 252 video recordings (averaging 18 seconds each), 135,000 video frames, and 270,000 force data points.

The \emph{Grasp} process is designed to study object manipulation dynamics. When resumed, it will involve systematic testing of grasp success/failure conditions and post-grasp stability (slip detection). Planned trials include successful grasps, failed attempts, stable holding phases, and controlled release maneuvers leading to slip events. All trials will maintain consistent data formatting with the touch process, featuring synchronized 720p video and sensor recordings.

\paragraph{Training Details} Table~\ref{tab:hyper_vtla_vta} list the training details for the models.

\subsection{More Results}

\begin{figure}[t!]
    \centering
    \includegraphics[width=1\linewidth]{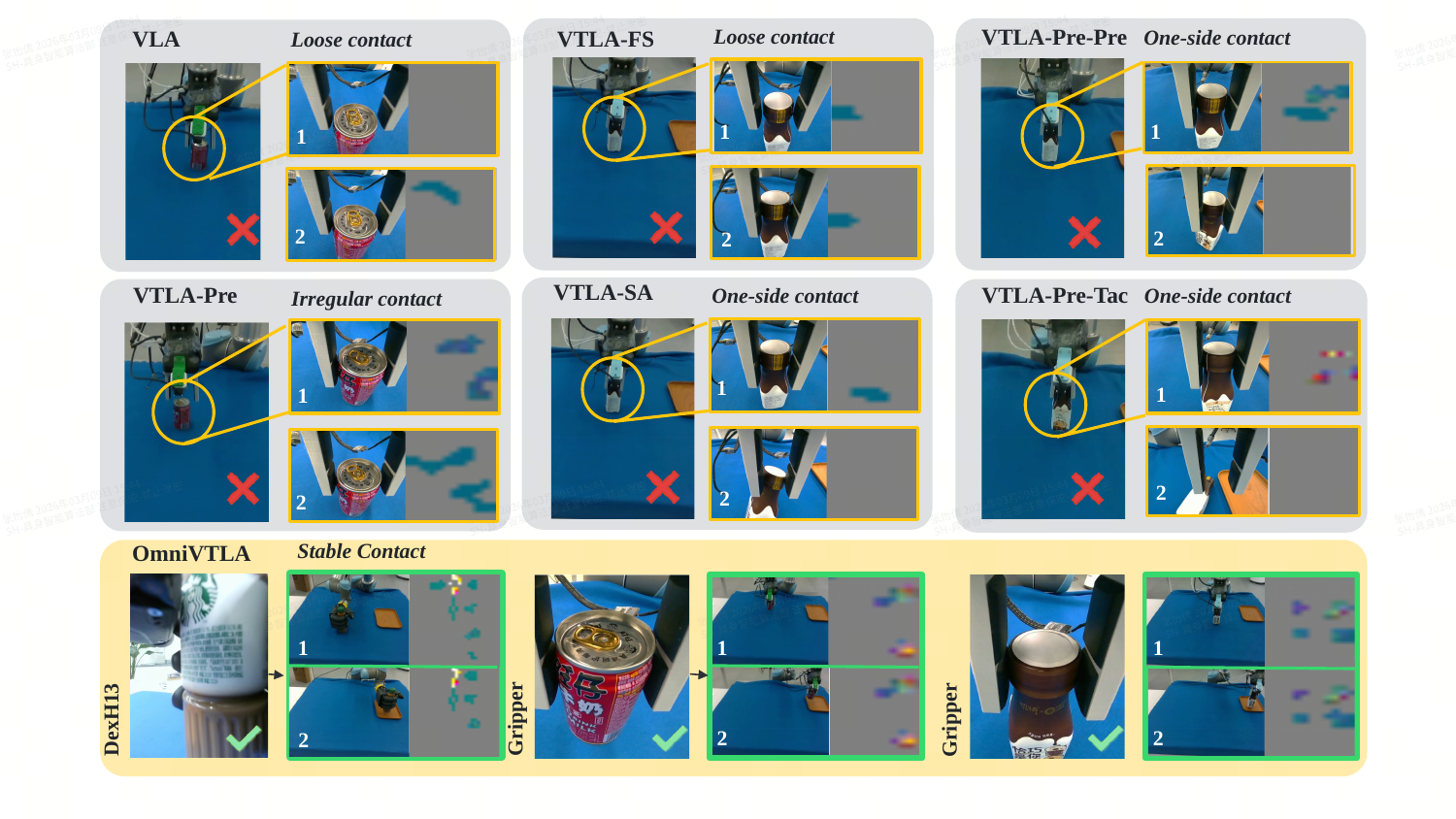}
    \caption{Pick and place visualization of several failed cases due to insufficient contact awareness, loose contact or irregular contact, and our proposed OmniVTLA achieves successful grasping and stable contact owing to full tactile sensing.}
    \label{fig:vis-all}
\end{figure}

\begin{figure}[t!]
    \centering
    \includegraphics[width=0.9\linewidth]{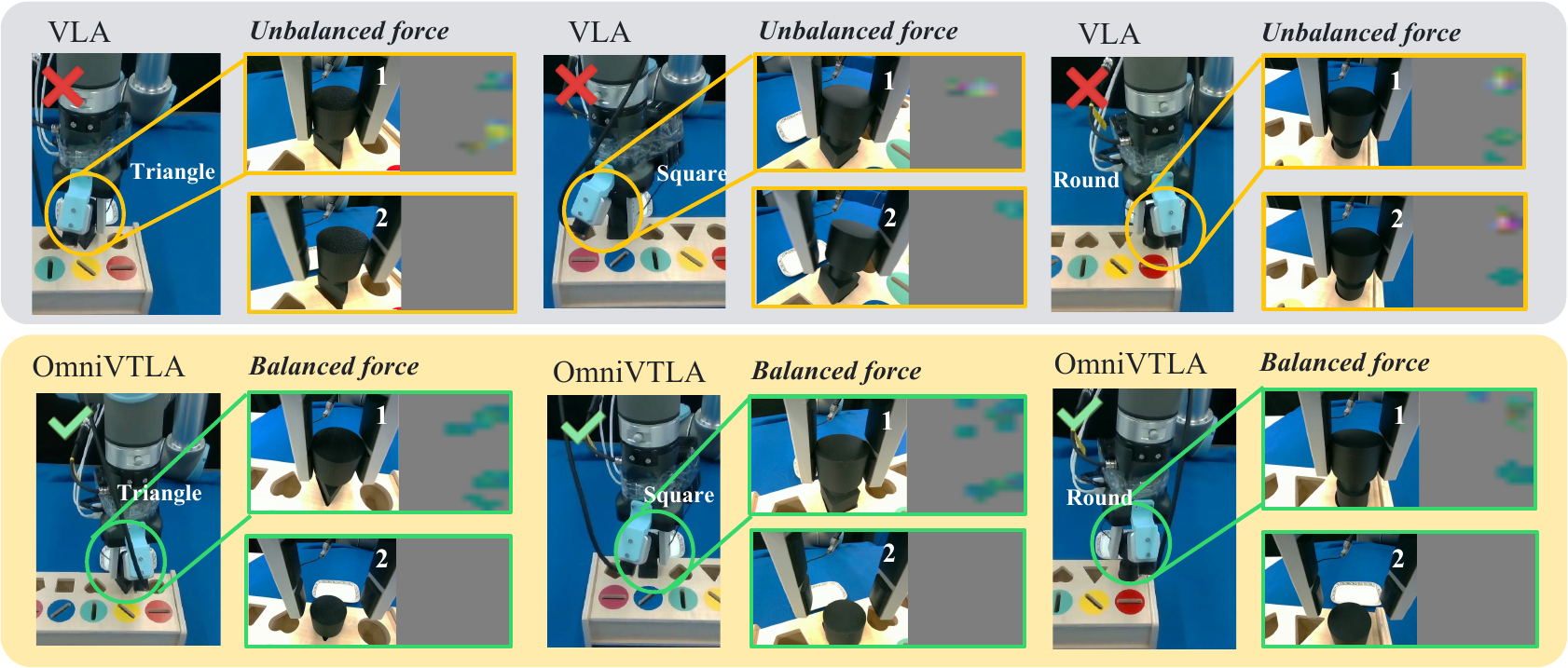}
    \caption{Peg insertion visualization of several failed cases for VLA due to unbalanced force, and some success cases of OmniVTLA owing to balanced forces.}
    \label{fig:vis-insertion}
\end{figure}

\paragraph{Qualitative Results} To understand the effectiveness of tactile sensing, we present some qualitative results for real-world experiments. The language prompt is ``\textit{Pick up the short can and move it to the plate}'', and we visualize the failed or successful cases for VLA, VTLA-Pre and OmniVTLA models (Fig.~\ref{fig:vis-all}). VLA models often fail to lift objects due to insufficient contact awareness, while VTLA-Pre struggles with persistent gripper adjustments without successful lifting. In contrast, OmniVTLA uses semantic tactile cues to stabilize grasps and execute smooth trajectories, as seen in successful lifts of the short can using the gripper and bottle using the dexterous hand. Visualizations of peg-insertion tasks for different models are drawn in Fig.~\ref{fig:vis-insertion}. OmniVTLA is able to insert the peg with balanced force to achieve higher SR.

\begin{figure}[t]
    \centering
    \includegraphics[width=0.8\linewidth]{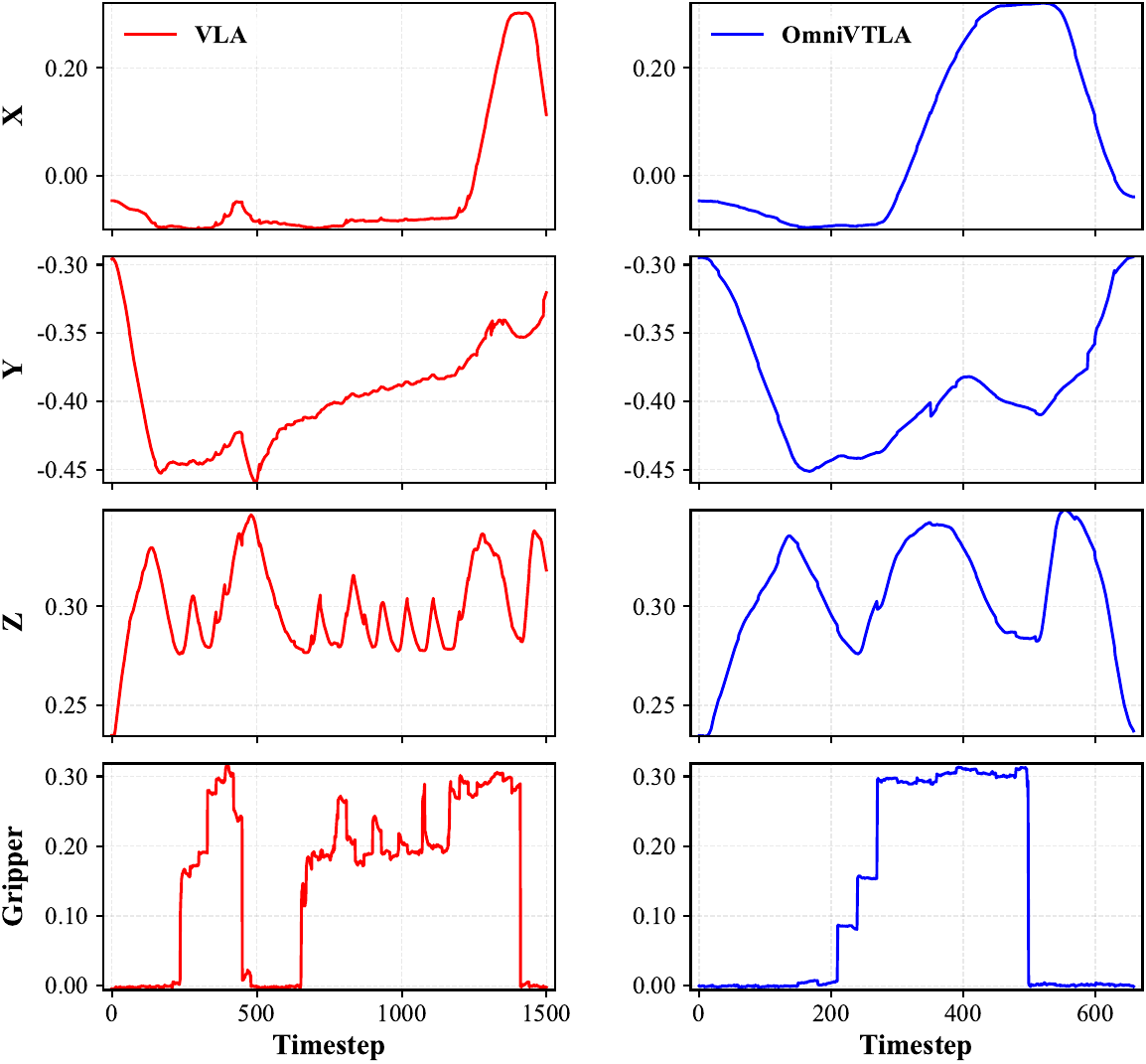}
    \caption{Action trajectories of OmniVTLA and VLA, where the first three figures show the spatial motion using the positions $(X, Y, Z)$ and the last figure shows the action of gripper closure.}
    \label{fig:trajectory}
\end{figure}

\paragraph{Comparison of Action Trajectories} 
Fig.~\ref{fig:trajectory} visualizes the smoothness of motion along the trajectory for two cases. The first three figures show the spatial motion using the positions $(X, Y, Z)$ and the last figure shows the action of gripper opening/closure. OmniVTLA exhibits superior motion smoothness through the entire process, and complete the task stably and successfully. In contrast, the trajectory of VLA is more erratic, with noticable jitter, instability and occasional dropping. 



%% file: paper.bib
@STRING{CVPR  = "IEEE Conf. Comput. Vis. Pattern Recog."}

@String(CVPR= {IEEE Conference on Computer Vision and Pattern Recognition (CVPR)})

@String(ICRA={IEEE international conference on robotics and automation (ICRA)})

@article{CLTP,
    title={CLTP: Contrastive Language-Tactile Pre-training for 3D Contact Geometry Understanding}, 
    author={Wenxuan Ma and Xiaoge Cao and Yixiang Zhang and Chaofan Zhang and Shaobo Yang and Peng Hao and Bin Fang and Yinghao Cai and Shaowei Cui and Shuo Wang},
  journal={arXiv preprint 2505.08194},
  year={2025}
}

@article{FG-CLTP,
  title={FG-CLTP: Fine-Grained Contrastive Language Tactile Pretraining for Robotic Manipulation},
  author={Wenxuan Ma and Chaofan Zhang and Yinghao Cai and Guocai Yao and Shaowei Cui and Shuo Wang},
  journal={arXiv preprint arXiv:2603.10871},
  year={2026}
}

@article{anytouch2,
  title={AnyTouch 2: General Optical Tactile Representation Learning For Dynamic Tactile Perception}, 
  author={Ruoxuan Feng and Yuxuan Zhou and Siyu Mei and Dongzhan Zhou and Pengwei Wang and Shaowei Cui and Bin Fang and Guocai Yao and Di Hu},
  journal={arXiv preprint arXiv:2602.09617},
  year={2026}
}

@article{octopi,
    title={Octopi: Object Property Reasoning with Large Tactile-Language Models}, 
    author={Samson Yu and Kelvin Lin and Anxing Xiao and Jiafei Duan and Harold Soh},
    journal={arXiv preprint arXiv:2405.02794},
    year={2024}
}

@ARTICLE{hall2006,
    author={E. Ramsden},
    title={Hall-Effect Sensor: Theory and Applications},
    year={2006},
    journal={Elsevier}
}

@article{dosovitskiy2020image,
  title={An image is worth 16x16 words: Transformers for image recognition at scale},
  author={Dosovitskiy, Alexey and Beyer, Lucas and Kolesnikov, Alexander and Weissenborn, Dirk and Zhai, Xiaohua and Unterthiner, Thomas and Dehghani, Mostafa and Minderer, Matthias and Heigold, Georg and Gelly, Sylvain and others},
  journal={arXiv preprint arXiv:2010.11929},
  year={2020}
}

@article{brohan2023rt,
  title={Rt-2: Vision-language-action models transfer web knowledge to robotic control},
  author={Brohan, Anthony and Brown, Noah and Carbajal, Justice and Chebotar, Yevgen and others},
  journal={arXiv preprint arXiv:2307.15818},
  year={2023}
}

@article{kim2024openvla,
  title={Openvla: An open-source vision-language-action model},
  author={Kim, Moo Jin and Pertsch, Karl and Karamcheti, Siddharth and Xiao, Ted and Balakrishna, Ashwin and Nair, Suraj and Rafailov, Rafael and Foster, Ethan and Lam, Grace and others},
  journal={arXiv preprint arXiv:2406.09246},
  year={2024}
}

@misc{black2024pi_0,
      title={ $\pi_0$: A Vision-Language-Action Flow Model for General Robot Control}, 
      author={Kevin Black and Noah Brown and Danny Driess and Adnan Esmail and Michael Equi and Chelsea Finn and Niccolo Fusai and Lachy Groom and Karol Hausman and Brian Ichter and others},
      year={2024},
      eprint={2410.24164},
      archivePrefix={arXiv},
      primaryClass={cs.LG},
}

@article{wen2025tinyvla,
  title={Tinyvla: Towards fast, data-efficient vision-language-action models for robotic manipulation},
  author={Wen, Junjie and others},
  journal={IEEE Robotics and Automation Letters},
  year={2025},
  publisher={IEEE}
}

@article{zhen20243d,
  title={3d-vla: A 3d vision-language-action generative world model},
  author={Zhen, Haoyu and Qiu, Xiaowen and Chen, Peihao and Yang, Jincheng and Yan, Xin and Du, Yilun and Hong, Yining and Gan, Chuang},
  journal={arXiv preprint arXiv:2403.09631},
  year={2024}
}

@article{liu2023visual,
  title={Visual instruction tuning},
  author={Liu, Haotian and Li, Chunyuan and others},
  journal={Advances in neural information processing systems},
  volume={36},
  pages={34892--34916},
  year={2023}
}

@article{li2024llava,
  title={Llava-onevision: Easy visual task transfer},
  author={Li, Bo and Zhang, Yuanhan and Guo, Dong and Zhang, Renrui and Li, Feng and Zhang, Hao and Zhang, Kaichen and Zhang, Peiyuan and Li, Yanwei and Liu, Ziwei and others},
  journal={arXiv preprint arXiv:2408.03326},
  year={2024}
}

@article{zhang2025videollama,
  title={VideoLLaMA 3: Frontier Multimodal Foundation Models for Image and Video Understanding},
  author={Zhang, Boqiang and Li, Kehan and Cheng, Zesen and Hu, Zhiqiang and Yuan, Yuqian and Chen, Guanzheng and Leng, Sicong and Jiang, Yuming and Zhang, Hang and Li, Xin and others},
  journal={arXiv preprint arXiv:2501.13106},
  year={2025}
}

@article{bai2025qwen2,
  title={Qwen2. 5-VL Technical Report},
  author={Bai, Shuai and Chen, Keqin and Liu, Xuejing and Wang, Jialin and others},
  journal={arXiv preprint arXiv:2502.13923},
  year={2025}
}

@inproceedings{radford2021learning,
  title={Learning transferable visual models from natural language supervision},
  author={Radford, Alec and Kim, Jong Wook and Hallacy, Chris and Ramesh, Aditya and Goh, Gabriel and Agarwal, Sandhini and Sastry, Girish and Askell, Amanda and Mishkin, Pamela and Clark, Jack and others},
  booktitle={International conference on machine learning},
  pages={8748--8763},
  year={2021},
  organization={PMLR}
}

@inproceedings{zhai2023sigmoid,
  title={Sigmoid loss for language image pre-training},
  author={Zhai, Xiaohua and Mustafa, Basil and Kolesnikov, Alexander and Beyer, Lucas},
  booktitle={Proceedings of the IEEE/CVF international conference on computer vision},
  pages={11975--11986},
  year={2023}
}

@article{chi2023diffusion,
  title={Diffusion policy: Visuomotor policy learning via action diffusion},
  author={Chi, Cheng and Xu, Zhenjia and Feng, Siyuan and Cousineau, Eric and Du, Yilun and Burchfiel, Benjamin and Tedrake, Russ and Song, Shuran},
  journal={The International Journal of Robotics Research},
  pages={02783649241273668},
  year={2023},
  publisher={SAGE Publications Sage UK: London, England}
}

@article{rotation6d,
  author       = {Yi Zhou and
                  Connelly Barnes and
                  Jingwan Lu and
                  others},
  title        = {On the Continuity of Rotation Representations in Neural Networks},
  journal      = {CoRR},
  volume       = {abs/1812.07035},
  year         = {2018},
  eprinttype    = {arXiv},
  eprint       = {1812.07035},
  bibsource    = {dblp computer science bibliography, https://dblp.org}
}

@article{rt2,
  title={Rt-2: Vision-language-action models transfer web knowledge to robotic control},
  author={Brohan, Anthony and Brown, Noah and Carbajal, Justice and Chebotar, Yevgen and Chen, Xi and Choromanski, Krzysztof and Ding, Tianli and Driess, Danny and Dubey, Avinava and Finn, Chelsea and others},
  journal={arXiv preprint arXiv:2307.15818},
  year={2023}
}

@article{openvla,
  title={Openvla: An open-source vision-language-action model},
  author={Kim, Moo Jin and Pertsch, Karl and Karamcheti, Siddharth and Xiao, Ted and Balakrishna, Ashwin and Nair, Suraj and Rafailov, Rafael and Foster, Ethan and Lam, Grace and others},
  journal={arXiv preprint arXiv:2406.09246},
  year={2024}
}

@article{rdt-1b,
  title={Rdt-1b: a diffusion foundation model for bimanual manipulation},
  author={Liu, Songming and Wu, Lingxuan and Li, Bangguo and Tan, Hengkai and Chen, Huayu and Wang, Zhengyi and Xu, Ke and Su, Hang and Zhu, Jun},
  journal={arXiv preprint arXiv:2410.07864},
  year={2024}
}

@article{beyond,
  title={Beyond sight: Finetuning generalist robot policies with heterogeneous sensors via language grounding},
  author={Jones, Joshua and Mees, Oier and Sferrazza, Carmelo and Stachowicz, Kyle and Abbeel, Pieter and Levine, Sergey},
  journal={arXiv preprint arXiv:2501.04693},
  year={2025}
}

@article{tla,
  title={Tla: Tactile-language-action model for contact-rich manipulation},
  author={Hao, Peng and Zhang, Chaofan and Li, Dingzhe and Cao, Xiaoge and Hao, Xiaoshuai and Cui, Shaowei and Wang, Shuo},
  journal={arXiv preprint arXiv:2503.08548},
  year={2025}
}

@article{vtla,
  title={VTLA: Vision-Tactile-Language-Action Model with Preference Learning for Insertion Manipulation},
  author={Zhang, Chaofan and Hao, Peng and Cao, Xiaoge and Hao, Xiaoshuai and Cui, Shaowei and Wang, Shuo},
  journal={arXiv preprint arXiv:2505.09577},
  year={2025}
}

@article{octo,
  title={Octo: An open-source generalist robot policy},
  author={Team, Octo Model and Ghosh, Dibya and Walke, Homer and Pertsch, Karl and Black, Kevin and Mees, Oier and Dasari, Sudeep and Hejna, Joey and Kreiman, Tobias and Xu, Charles and others},
  journal={arXiv preprint arXiv:2405.12213},
  year={2024}
}

@article{gr00t,
  title={Gr00t n1: An open foundation model for generalist humanoid robots},
  author={Bjorck, Johan and Casta{\~n}eda, Fernando and Cherniadev, Nikita and Da, Xingye and Ding, Runyu and Fan, Linxi and Fang, Yu and Fox, Dieter and Hu, Fengyuan and Huang, Spencer and others},
  journal={arXiv preprint arXiv:2503.14734},
  year={2025}
}

@article{anytouch,
  title={Anytouch: Learning unified static-dynamic representation across multiple visuo-tactile sensors},
  author={Feng, Ruoxuan and Hu, Jiangyu and Xia, Wenke and Gao, Tianci and Shen, Ao and Sun, Yuhao and Fang, Bin and Hu, Di},
  journal={arXiv preprint arXiv:2502.12191},
  year={2025}
}

@inproceedings{unitouch,
  title={Binding touch to everything: Learning unified multimodal tactile representations},
  author={Yang, Fengyu and Feng, Chao and Chen, Ziyang and Park, Hyoungseob and Wang, Daniel and Dou, Yiming and Zeng, Ziyao and Chen, Xien and Gangopadhyay, Rit and Owens, Andrew and others},
  booktitle={Proceedings of the IEEE/CVF Conference on Computer Vision and Pattern Recognition},
  pages={26340--26353},
  year={2024}
}

@article{t3,
  title={Transferable tactile transformers for representation learning across diverse sensors and tasks},
  author={Zhao, Jialiang and Ma, Yuxiang and Wang, Lirui and Adelson, Edward H},
  journal={arXiv preprint arXiv:2406.13640},
  year={2024}
}

@article{tvl,
  title={A touch, vision, and language dataset for multimodal alignment},
  author={Fu, Letian and Datta, Gaurav and Huang, Huang and Panitch, William Chung-Ho and Drake, Jaimyn and Ortiz, Joseph and Mukadam, Mustafa and Lambeta, Mike and Calandra, Roberto and Goldberg, Ken},
  journal={arXiv preprint arXiv:2402.13232},
  year={2024}
}

@article{cheng2025touch100k,
  title={Touch100k: A large-scale touch-language-vision dataset for touch-centric multimodal representation},
  author={Cheng, Ning and Xu, Jinan and Guan, Changhao and others},
  journal={Information Fusion},
  pages={103305},
  year={2025},
  publisher={Elsevier}
}

@misc{PaxiniSensor,
      title={PX-6AX: ITPU Tactile Processing Unit}, 
      author={Paxini},
      year={2025},
      url={https://paxini.com/ax/gen2}, 
}

@inproceedings{cui2020grasp,
  title={Grasp state assessment of deformable objects using visual-tactile fusion perception},
  author={Cui, Shaowei and Wang, Rui and others},
  booktitle={International Conference on Robotics and Automation (ICRA)},
  pages={538--544},
  year={2020},
  organization={IEEE}
}

@inproceedings{li2018slip,
  title={Slip detection with combined tactile and visual information},
  author={Li, Jianhua and Dong, Siyuan and Adelson, Edward},
  booktitle={2018 IEEE International Conference on Robotics and Automation (ICRA)},
  pages={7772--7777},
  year={2018},
  organization={IEEE}
}

@inproceedings{hansen2022visuotactile,
  title={Visuotactile-rl: Learning multimodal manipulation policies with deep reinforcement learning},
  author={Hansen, Johanna and Hogan, Francois and Rivkin, Dmitriy and Meger, David and Jenkin, Michael and Dudek, Gregory},
  booktitle={2022 International Conference on Robotics and Automation (ICRA)},
  pages={8298--8304},
  year={2022},
  organization={IEEE}
}

@article{lee2020making,
  title={Making sense of vision and touch: Learning multimodal representations for contact-rich tasks},
  author={Lee, Michelle A and Zhu, Yuke and Zachares, Peter and Tan, Matthew and Srinivasan, Krishnan and Savarese, Silvio and Fei-Fei, Li and Garg, Animesh and Bohg, Jeannette},
  journal={IEEE Transactions on Robotics},
  volume={36},
  number={3},
  pages={582--596},
  year={2020},
  publisher={IEEE}
}

@article{liu2025vitamin,
  title={Vitamin: Learning contact-rich tasks through robot-free visuo-tactile manipulation interface},
  author={Liu, Fangchen and Li, Chuanyu and Qin, Yihua and Shaw, Ankit and Xu, Jing and Abbeel, Pieter and Chen, Rui},
  journal={arXiv preprint arXiv:2504.06156},
  year={2025}
}

@article{xue2025reactive,
  title={Reactive diffusion policy: Slow-fast visual-tactile policy learning for contact-rich manipulation},
  author={Xue, Han and Ren, Jieji and Chen, Wendi and Zhang, Gu and Fang, Yuan and Gu, Guoying and Xu, Huazhe and Lu, Cewu},
  journal={arXiv preprint arXiv:2503.02881},
  year={2025}
}

@article{huang20243d,
  title={3D-ViTac: Learning Fine-Grained Manipulation with Visuo-Tactile Sensing},
  author={Huang, Binghao and Wang, Yixuan and Yang, Xinyi and Luo, Yiyue and Li, Yunzhu},
  journal={arXiv preprint arXiv:2410.24091},
  year={2024}
}

@article{yu2023mimictouch,
  title={MimicTouch: Leveraging Multi-modal Human Tactile Demonstrations for Contact-rich Manipulation},
  author={Yu, Kelin and Han, Yunhai and Wang, Qixian and Saxena, Vaibhav and Xu, Danfei and Zhao, Ye},
  journal={arXiv preprint arXiv:2310.16917},
  year={2023}
}

@article{lin2024learning,
  title={Learning visuotactile skills with two multifingered hands},
  author={Lin, Toru and Zhang, Yu and others},
  journal={arXiv preprint arXiv:2404.16823},
  year={2024}
}

@article{calandra2018more,
  title={More than a feeling: Learning to grasp and regrasp using vision and touch},
  author={Calandra, Roberto and Owens, Andrew and Jayaraman, Dinesh and Lin, Justin and Yuan, Wenzhen and Malik, Jitendra and Adelson, Edward H and Levine, Sergey},
  journal={IEEE Robotics and Automation Letters},
  volume={3},
  number={4},
  pages={3300--3307},
  year={2018},
  publisher={IEEE}
}

@article{team2025gemini,
  title={Gemini robotics: Bringing ai into the physical world},
  author={Team, Gemini Robotics and Abeyruwan, Saminda and Ainslie, Joshua and others},
  journal={arXiv preprint, arXiv:2503.20020},
  year={2025}
}

@article{qu2025spatialvla,
  title={SpatialVLA: Exploring Spatial Representations for Visual-Language-Action Model},
  author={Qu, Delin and Song, Haoming and Chen, Qizhi and Yao, Yuanqi and Ye, Xinyi and Ding, Yan and Wang, Zhigang and Gu, JiaYuan and Zhao, Bin and others},
  journal={arXiv preprint arXiv:2501.15830},
  year={2025}
}

@article{shukor2025smolvla,
  title={SmolVLA: A Vision-Language-Action Model for Affordable and Efficient Robotics},
  author={Shukor, Mustafa and Aubakirova, Dana and others},
  journal={arXiv preprint arXiv:2506.01844},
  year={2025}
}

@article{lin2025onetwovla,
  title={OneTwoVLA: A Unified Vision-Language-Action Model with Adaptive Reasoning},
  author={Lin, Fanqi and Nai, Ruiqian and Hu, Yingdong and You, Jiacheng and Zhao, Junming and Gao, Yang},
  journal={arXiv preprint arXiv:2505.11917},
  year={2025}
}

@inproceedings{zhao2025cot,
  title={Cot-VLA: Visual chain-of-thought reasoning for vision-language-action models},
  author={Zhao, Qingqing and Lu, Yao and others},
  booktitle={CVPR},
  pages={1702--1713},
  year={2025}
}

@article{huang2025tactile,
  title={Tactile-VLA: Unlocking Vision-Language-Action Model's Physical Knowledge for Tactile Generalization},
  author={Huang, Jialei and Wang, Shuo and Lin, Fanqi and Hu, Yihang and Wen, Chuan and Gao, Yang},
  journal={arXiv preprint arXiv:2507.09160},
  year={2025}
}

@article{yu2025forcevla,
  title={ForceVLA: Enhancing VLA Models with a Force-aware MoE for Contact-rich Manipulation},
  author={Yu, Jiawen others},
  journal={arXiv preprint arXiv:2505.22159},
  year={2025}
}

@article{hu2025dexterous,
  title={Dexterous in-hand manipulation of slender cylindrical objects through deep reinforcement learning with tactile sensing},
  author={Hu, Wenbin and Huang, Bidan and Lee, Wang Wei and Yang, Sicheng and Zheng, Yu and Li, Zhibin},
  journal={Robotics and Autonomous Systems},
  volume={186},
  pages={104904},
  year={2025},
  publisher={Elsevier}
}
